\documentclass{article} 
\usepackage{iclr2027_conference,times}

\usepackage{amsmath,amsfonts,bm}

\def\eqref#1{equation~\ref{#1}}

\def\1{\bm{1}}

\DeclareMathAlphabet{\mathsfit}{\encodingdefault}{\sfdefault}{m}{sl}
\SetMathAlphabet{\mathsfit}{bold}{\encodingdefault}{\sfdefault}{bx}{n}

\usepackage{hyperref}
\usepackage{url}
\usepackage{booktabs}
\usepackage{multirow}
\usepackage{multicol}
\usepackage{amsmath}
\usepackage{amssymb}
\usepackage{mathtools}
\usepackage{amsthm}

\usepackage{algorithm}
\usepackage{algorithmic}
\usepackage{wrapfig}
\usepackage{subcaption}

\title{MatGPTQ: Efficient and Accurate Inference over Nested Quantized Models}

\author{
Maximilian Kleinegger\thanks{Work done during an internship at ISTA.} \\
TU Wien \& ISTA
\And
Elvir Crnčević \\
Red Hat AI
\And
Dan Alistarh\thanks{This author was affiliated with Red Hat AI during part of this work.}\thanks{Correspondence to: \texttt{dan.alistarh@ist.ac.at}.} \\
ISTA
}
\iclrfinalcopy 
\begin{document}

\maketitle

\begin{abstract}
Matryoshka Quantization (MatQuant), Any-Precision-LLM (AP) and AnyBCQ (AB) are recent quantization approaches showing that a single integer-quantized model can be served across multiple precisions. In this paradigm, lower-precision models are extracted from a higher-precision model by simply reading fewer bits of the weights. This enables a single checkpoint to cover a wide range of memory and latency budgets, but makes both quantization and efficient execution substantially harder. Existing methods rely on expensive quantization-aware training (QAT) or gradient-based post-training quantization (PTQ) rather than fast one-shot PTQ, and offer limited system support: dedicated kernels are either missing or restricted to single- or small-batch decoding.
We address these limitations with \emph{Post-Training Matryoshka Quantization} (MatGPTQ), an end-to-end pipeline for nested-model quantization and inference. MatGPTQ casts Matryoshka quantization as multi-precision error compensation, producing a single ``sliceable'' parent model jointly optimized for multiple target precisions in one pass over a small calibration set. We further refine the MatQuant representation so that an $r$-bit model reads exactly $r$ bits, and introduce the first dedicated inference kernels for this format, supporting batch sizes beyond one and integrated into vLLM. Across standard LLMs and benchmarks, MatGPTQ outperforms MatQuant while remaining competitive with AP and AB at the smallest checkpoint size, and our kernels achieve end-to-end speedups of up to 3.5$\times$ over BF16 at the low-bit regime. Overall, MatGPTQ makes nested quantized models practical to serve from a single, compact checkpoint. Code is available at \url{https://github.com/IST-DASLab/MatGPTQ}.
\end{abstract}

\section{Introduction}
Quantization is currently the most practical way to shrink the memory and bandwidth costs of large language models (LLMs): by mapping full-precision weights to low-bit representations, one can enable single‑GPU or edge deployment, while mostly retaining model accuracy. One‑shot, post‑training quantization (PTQ) is attractive for its low cost and strong accuracy in the 3--4 bit regime~\citep{gptq, kurtic2024give}, whereas optimization‑based methods such as QAT and block‑wise, gradient‑calibrated PTQ (e.g., OmniQuant~\citep{shao2023omniquant}) achieve higher accuracy in the low-bit regime (i.e: 2--3 bits), but at the expense of extra computational cost.

Recently, \emph{Matryoshka Quantization} (MatQuant)~\citep{matryoshka_quantization}, \emph{Any-Precision-LLM} (AP)~\citep{park2024any} and \emph{AnyBCQ} (AB)~\citep{ICLR2026_ca9eaef0} provided a fresh approach to model compression by turning integer bit representations into a \textit{nested representation}: a single checkpoint supports several inference precisions. Even though MatQuant, AP and AB pursue this goal, they achieve it differently. MatQuant jointly optimizes a high-precision parent representation whose prefixes define lower-precision models, whereas AP and AB construct the hierarchy progressively from a low-precision model. Therefore, these approaches trade off quantization cost, representation complexity and hardware execution in different ways. 

Despite their attractive storage flexibility, existing nested representations remain difficult to deploy efficiently. MatQuant relies on a single shared-scale integer representation, but is reliant on gradient-based methods such as QAT and OmniQuant. Furthermore, MatQuant’s lower-precision models are not \emph{true-to-size}, since extracting an $r$-bit model requires reading $r+1$ bits from the parent representation. We refer to this operation as \emph{slicing} throughout the paper. The original work also did not provide open-source code or dedicated kernels for the format, making efficient deployment difficult. On the other hand, AP and AB kernels have limitations: AP kernels only support small batch sizes $\leq 16$ during decoding, while AB kernels are only available for single-batch decoding. For prefill and higher-batch decoding, both methods fall back to dequantization followed by cuBLAS. In addition, both methods rely on more complex and overall more memory-heavy representations, with AP requiring bit-width-specific centroid lookup tables and AB requiring individual scales for each supported bit-width. Consequently, there remains a gap between storing many precision levels efficiently and executing them efficiently across realistic workloads.

\textbf{Contributions.} We revisit the MatQuant representation from a deployment perspective and ask whether its nested, shared-scale integer structure can retain the accuracy of recent multi-precision approaches while admitting efficient end-to-end execution across prefill, decoding, and batch sizes beyond one. We answer this with \emph{MatGPTQ}, an end-to-end framework that combines one-shot multi-precision PTQ with a deployment-friendly nested representation and dedicated GPU kernels. In summary, our contributions are:

\begin{enumerate}
    \item \textbf{Comparative view of Nested Quantization}. We study MatQuant, AP, and AB as alternative realizations of single-checkpoint multi-precision inference and characterize their trade-offs in construction, representation flexibility and hardware execution.
    \item \textbf{Multi-precision One-Shot PTQ}. We adapt GPTQ to jointly optimize a single parent model across multiple target precisions using cross-bit error compensation.
    \item \textbf{Deployment-friendly Representation.} We introduce true-prefix slicing, improving over \emph{MatQuant}, while keeping the single shared scale format. Furthermore, we provide GPU kernels for nested inference at 2, 3, 4, 6 and 8 bits, supporting prefill, decoding, and larger batch sizes, together with efficient packing and vLLM integration~\citep{vllm}.
    \item \textbf{Empirical Validation}. We evaluate MatGPTQ on Llama3.1-8B-Instruct and Qwen3-8B against independently quantized GPTQ, MatQuant, AP, and AB. MatGPTQ substantially improves over MatQuant, remains competitive at every precision with AP and AB, while using the least complex representation.
    
\end{enumerate}

\section{Background \& Related Work}
\label{sec:background}

\subsection{Post-training Quantization}
\label{sec:ptq}

\textbf{Quantization} is often applied as a \emph{post-training (PTQ)} step, in one-shot to already trained models. We focus on such one-shot PTQ methods in this paper, by contrast to more expensive quantization-aware training (QAT). 
For PTQ, one can differentiate between \emph{data-free} approaches, which use simpler round-to-nearest (RTN) projections~\citep{dettmers20168bit, badri2023hqq}, but tend to drop accuracy even at 4-bits, and \emph{data-aware} approaches such as GPTQ \citep{gptq} or AWQ~\citep{Lin2024_AWQ}, which utilize a calibration set that is passed through the model during quantization. Specifically, a common approach in PTQ is to quantize each linear layer $\ell$ with weights $W_\ell$, in linear order, to produce the quantized weights $Q_\ell$. 
Please see Section~\ref{sec:matgptq} for an overview of the MatGPTQ algorithm.

\subsection{Nested multi-precision quantization}
\label{sec:nested-quant}

We call models nested when higher bit-width models ``contain'' the lower-bit-width models, within a single checkpoint.

\textbf{Matryoshka Quantization} (MatQuant)~\citep{matryoshka_quantization} is a shared-scale framework for producing nested models: from a $c$-bit master code $q$, an $r$-bit model ($0 < r \leq c$) is obtained by \emph{slicing},
\begin{equation}
\label{eq:slicing}
S(q, r) = \mathrm{clamp}\!\left(\mathrm{round}\!\left(\frac{q}{2^{c-r}}\right),\, 0,\, 2^{r}-1\right) \cdot 2^{c-r}.
\end{equation}
Here, the right shift keeps the $r$ most significant bits, while the left shift maps the value back into the $c$-bit code space, allowing the same scale and zero-point to be reused while approximating the original $c$-bit value. Truncation, however, introduces a systematic underestimation and therefore loses information. To reduce this effect, MatQuant uses the $(r+1)$-th bit for rounding, which can overflow the valid $r$-bit range and therefore requires an additional clamp operation. Consequently, extracting an $r$-bit model requires reading $r+1$ bits per weight together with additional rounding and clamping operations, complicating efficient deployment. See Appendix~\ref{sec:improved_slicing} for more details.

\citet{matryoshka_quantization} then define their objective function, which minimizes over $\theta$, representing the set of model and auxiliary parameters being optimized. The loss function $\mathcal{L}(\cdot)$ is evaluated for a fixed target set of bit-widths $R = \{r_1, r_2, ..., r_K\}$, under the quantization method $Q(\cdot)$ and forward pass $F(\cdot)$, and is calculated as:
\begin{equation}
\label{eq:matquant_obj}
\min_{\theta}
\frac{1}{N}\sum_{i=1}^{N}\sum_{r \in R}
\lambda_r \,
\mathcal{L}\!\left(F(S(Q(\theta,c),r)); \mathcal{D}_i\right),
\end{equation}
Here, $\mathcal{D}_i$ represents either the input-label pair used for QAT or
the layer inputs and full-precision outputs used for OmniQuant. The parameter $\lambda_r$ is used to weight the importance of individual bit-widths, allowing the optimization to emphasize different bit-widths accordingly.  Overall, the MatQuant representation follows the default quantization representation, which is a combination of quantized weights, scales and zero-points that are grouped for better performance.

Relatedly, TruncQuant~\citep{kim2025truncquant} enables variable-precision inference through truncation of low-order bits, but relies on truncation-aware training rather than one-shot PTQ and does not target the nested LLM deployment setting considered here.

\textbf{Any-Precision-LLM} (AP)~\citep{park2024any} introduces an
\emph{up-scaling} method based on SqueezeLLM's~\citep{kim2023squeezellm}
clustered, non-uniform quantization. AP starts by constructing the lowest-bit \emph{seed} model using the corresponding centroid-based representation. Afterward, additional bits and therefore centroids are progressively added to increase the number of representable centroids and construct higher-bit-width models. Generally, the centroid values are stored within lookup tables, while the quantized weights store indices pointing to the respective centroid values.

The corresponding quantized weights are stored using bit-planes, with each additional bit represented by a separate plane. At inference time, an $r$-bit model is obtained by loading the first $r$ bit-planes and combining them to recover the corresponding centroid indices, which are then used to look up the respective centroid values. This introduces additional representation complexity, as AP relies on centroid lookup tables, which also need to be stored for each supported bit-width and layer, to reconstruct the quantized weights. They choose this approach since their up-scaling method cannot be directly applied to GPTQ or AWQ.

\textbf{AnyBCQ} (AB)~\citep{ICLR2026_ca9eaef0} builds on Binary-Coded Quantization (BCQ). AB starts by constructing the lowest-bit \emph{base} model using BCQ. Afterward, higher bit-widths are progressively constructed by freezing the previously obtained binary bases and adding a new binary basis together with its corresponding scale. The newly added basis is initialized from the residual error of the lower-bit model and subsequently refined, allowing each higher bit-width to reuse all previously constructed binary bases.

The corresponding binary bases are stored as individual bit-planes. More precisely, for a bit-width $r$, the quantized weight matrix $\widehat{\mathbf{W}}_r$ is represented as a linear combination of $r$ binary bases and their corresponding real-valued scales, $\widehat{\mathbf{W}}_r = \sum_{i=1}^{r} \alpha_i \mathbf{B}_i$, where each $\mathbf{B}_i$ is a binary matrix with $\{-1,1\}$ entries and $\alpha_i \in \mathbb{R}$ represents its corresponding scale. At inference time, an $r$-bit model is obtained by loading the first $r$ binary bit-planes together with their respective scales and combining them to reconstruct the quantized weights. Unlike AP, this representation does not require centroid lookup tables and instead allows computation directly on the binary bit-planes. However, each additional bit introduces an additional scale and potentially an additional zero-point for each group, resulting in multiple quantization parameters being required for the nested representation.

\textbf{Further Related Work.} $\textnormal{D}^{2}\textnormal{MoE}$~\citep{wang2025d} also introduces a nested multi-precision representation, but specifically targets Mixture-of-Experts (MoE) layers and expert-wise execution. NestQuant~\citep{xie.tmc2025nestquant} also studies nested integer representations, but focuses on on-device DNN deployment and constructs nested models through a high-/low-bit decomposition. Therefore, we do not include these methods in our comparison.

\subsection{Comparison of nested representations}

\begin{table}[h!]
\centering
\caption{Comparison of multi-precision quantization methods.}
\label{tab:method_comparison}
\begin{tabular}{lcccc}
\toprule
\textbf{Property} & \textbf{MatQuant} & \textbf{AP} & \textbf{AB} & \textbf{MatGPTQ} \\
\midrule
Uses gradients during quantization & \checkmark &  $\sim$  & \checkmark & $\times$ \\
One-shot multi-precision PTQ & $\times$   & $\times$   & $\times$   & \checkmark \\
Single multi-precision checkpoint & \checkmark & \checkmark & \checkmark & \checkmark \\
Attention+FFN quantization & $\sim$ & \checkmark & \checkmark & \checkmark \\
Exact $r$-bit loading & $\times$ & \checkmark & \checkmark & \checkmark \\
Inference kernel & $\times$   & $\sim$ & $\sim$ & \checkmark \\
Open-source implementation & $\times$ & \checkmark & \checkmark & \checkmark \\
\midrule \midrule
Nested checkpoint size (Llama-3.1-8B)   & 8 GB     & 10 GB     & 13 GB     & 8 GB \\
Quantization time (Llama-3.1-8B, A6000) & $\sim7\,h$ & $\sim35\,min$ & $\sim18\,h$ & $\sim35\,min$ \\
\bottomrule
\end{tabular}
\end{table}

Next, we compare key characteristics of MatQuant, AP, and AB, as summarized in Table~\ref{tab:method_comparison}. First, all three methods rely on gradient information, which increases the computational cost and complexity of quantization and motivates the question of whether comparable performance can be achieved without gradients. However, they differ in when gradients are required: AP uses gradients only prior to quantization, whereas MatQuant and AB rely on them during the quantization process itself. Second, all methods provide a single checkpoint. 
However, MatQuant was only evaluated using FFN quantization for its PTQ experiments. Furthermore, MatQuant's representation does not allow exact $r$-bit loading, as an additional bit, for rounding, needs to be read. MatQuant does not provide any open-source implementation or kernel support, making adaptation and usage difficult.  AP and AB, on the other hand, quantize Attention+FFN while providing exact $r$-bit loading and their implementations are open-sourced, but they only provide limited kernel support. Specifically, AP only provides GEMM support for low batch decoding scenarios, while AB provides SIMT kernels for batch-size-1 decoding. Both rely on dequantization followed by cuBLAS for prefill and larger batch sizes. This represents a \emph{limitation} and leaves this as an open problem. Lastly, the complexity of the different representations can be observed in the storage overhead for the same model, as seen in Table~\ref{tab:method_comparison}. Here, MatQuant's simple format results in the smallest model-size, while especially AB requires substantially more storage, making deployment on memory-constrained devices harder.

In this context, our work presents the first post-training quantization (PTQ) and open-source variant of an improved MatQuant representation, enabling nested quantization of both Attention and FFN layers from a single checkpoint. We evaluate accuracy at 3, 4, 6 and 8 bits, while our kernels additionally support 2-bit execution. We introduce exact $r$-bit loading from a single checkpoint while retaining MatQuant's compact shared-scale representation. MatGPTQ achieves performance that is on par with independently quantized, non-nested PTQ models across the supported bit-widths. Furthermore, in direct comparison, it substantially improves over MatQuant's OmniQuant implementation and is competitive with or exceeds AP and AB at every precision. Finally, we provide dedicated CUDA kernels that enable efficient nested inference for both prefill and decoding, including batch sizes larger than one, together with the corresponding memory savings.

\section{Matryoshka GPTQ}
\label{sec:matgptq}
Since the original MatQuant construction is gradient-based and its slicing operation does not allow exact $r$-bit loading, we address both limitations from a PTQ and deployment perspective. We first simplify the representation for efficient system execution, and then adapt GPTQ to enable one-shot, full-model multi-precision quantization.

\subsection{Improved Representation}

Recall that the original slicing operator $S(\cdot, r)$ from Equation~\ref{eq:slicing} extracts the $r$ most significant bits using a right shift, followed by a left shift to push values back into the $c$-bit regime. However, it additionally requires reading the $(r{+}1)$-th bit for rounding, and a clamp operation to ensure that the rounded value remains within the valid $r$-bit range, which increases operation and memory complexity. To overcome these limitations, we introduce an exact $r$-bit slicing operation:
\begin{equation}
\label{eq:matgptq_slicing}
S'(q,r) = \left\lfloor \frac{q}{2^{c-r}} \right\rfloor \cdot 2^{c-r},\quad q \in \{0, \dots, 2^c-1\},
\end{equation}
which simply discards the lowest $d = c - r$ bits by performing only the originally intended right followed by left shift. Each slice is therefore a plain bit prefix of the master code: no additional bit is read, no carry can occur, and no clamping is required. 
Since truncation maps all codes of a cell to its lowest value, we recenter by $\delta_r = (2^d-1)/2$, which minimizes the root-mean-squared error for uniformly distributed codes. This matches the error of MatQuant's rounding while retaining exact $r$-bit loading. Additionally, at inference, $\delta_r$ is folded into the zero-point. Details are in  Appendix~\ref{sec:improved_slicing}.

\subsection{The MatGPTQ Algorithm}
For each layer $\ell$ with weights $W_\ell$, we optimize a set of target bit-widths $R=\{r_1,\dots,r_K\}\subseteq\{2,\dots,c\}$, where $c\in \mathbb{N}$ is the highest \emph{master} bit-width. This master representation is used as the basis for the shared scale and zero-point, while the lower bit-widths are derived from it. Recall that GPTQ solves the data-aware layer-wise quantization problem in Equation~\ref{eq:min_quant_obj} w.r.t. $Q_\ell$, for fixed $W_\ell$, calibration data $X_\ell$, and group-wise scale $s_\ell$ and zero-point $z_\ell$:
\begin{equation}
    \label{eq:min_quant_obj}
    \arg\min_{Q_\ell} \|\widehat{W}_\ell X_\ell - W_\ell X_\ell\|^2_2,\quad \widehat{W}_\ell = s_\ell\,(Q_\ell - z_\ell).
\end{equation}
We now adapt this layer-wise GPTQ objective to the MatQuant formulation of Equation~\ref{eq:matquant_obj}. Given $W_\ell$ and $X_\ell$, GPTQ leverages the layer-wise Hessian $H_\ell = 2X_\ell X_\ell^\top$ to quantize weights, for a single precision, in a fixed order while compensating for the error through updates to the remaining unquantized weights. Instead, we replace this single-precision error with the $\lambda$-weighted error across all target bit-widths, so that one set of master codes $Q^c_\ell$ is chosen for every precision at once:
\begin{equation}
\label{eq:min_matgpt_obj}
\arg\min_{Q^c_\ell,\,s_\ell,\,z_\ell}\;
\sum_{r \in R} \lambda_r \,
\bigl\| \widehat{W}_{\ell,r}\, X_\ell - W_\ell X_\ell \bigr\|_2^2,
\quad \widehat{W}_{\ell,r} = s_\ell\,\bigl(S'(Q^c_\ell, r) + \delta_r - z_\ell\bigr),
\end{equation}
where $S'$ is the exact slicing operator of Equation~\ref{eq:matgptq_slicing} and $\delta_r$ its truncation offset.

\textbf{From GPTQ to MatGPTQ.} This formulation changes GPTQ from optimizing a single target precision to jointly selecting one master representation across all $r \in R$. Realizing this objective requires three corresponding modifications to the original GPTQ algorithm:
\begin{enumerate}
    \item The group-wise scale and zero-point are no longer fitted against the master bit-width alone: we approximate MSE-optimal values via a grid search that evaluates each candidate pair $(s_\ell, z_\ell)$ against every $r \in R$. The algorithm can be found in Appendix~\ref{section:perf-alg2}.
    \item The master codes $Q^c_\ell$ are not obtained by rounding to the nearest value on the quantization grid, since the nearest $c$-bit code may have poor lower-bit prefixes. Instead, each code $q$ for a weight $w$ is selected by minimizing the cross-bit-width weighted error of Equation~\ref{eq:min_matgpt_obj} over its prefixes:
    \begin{equation}
    \label{eq:matgptq_round}
    q^\star = \arg\min_{q\in\{0,\dots,2^c-1\}}\;
    \sum_{r\in R}\lambda_r\,\bigl(s_\ell\,(S'(q,r)+\delta_r-z_\ell) - w\bigr)^2
    \end{equation}
    The algorithm can be found in Appendix~\ref{section:perf-alg2}, together with its memory and runtime cost.
    \item Error propagation must account for all bit-widths at once. Instead of propagating the single error that one bit-width would produce, as in native GPTQ, we propagate the $\lambda$-weighted average of the errors produced by all target bit-widths. 
    Since all bit-widths share the remaining weights and the same Hessian, this is the OBS update minimizing Equation~\ref{eq:min_matgpt_obj} over the remaining weights (for normalized $\lambda_r$).
    
\end{enumerate}

\textbf{Floating-Point.} Extending the MatQuant framework to floating-point representations remains challenging: the discarded low-order bits then belong to the mantissa or exponent and contribute to the value multiplicatively rather than additively~\citep{matryoshka_quantization}, so truncation no longer introduces a constant offset $\delta_r$ that can be folded into the zero-point. We leave this as a future direction.

\section{Fast Matryoshka Kernels}
\label{sec:kernel}

We implement a CUDA kernel that serves the nested bit-widths $r \in \{2,3,4,6,8\}$ from a single packed model, dequantizing the weights on the fly. It targets low-batch latency, uses only \texttt{mma.sync}, \texttt{cp.async} and \texttt{ldmatrix}, and therefore runs on Ampere, Ada and Hopper. It could be extended for Hopper-specific features, similarly to how  Marlin~\citep{frantar2024marlin} was extended to Machete~\citep{wilkinson2024machete} for Hopper. Given dequantized weights $W_D$ and FP16 or BF16 inputs $\mathbf{X} \in \mathbb{R}^{b \times d_\text{in}}$, it computes $\mathbf{Y} = \mathbf{X} W_D$, accumulating in FP32.

The implementation has three main components: First, we utilize the \texttt{mma.sync.m16n8k16} tensor-core instruction for computation. Since the batch $b$ is small, we compute $\mathbf{Y} = (W_D^\top \mathbf{X}^\top)^\top$ internally, so that the weights fill the 16-row operand and the tokens the 8-column operand, while the layer keeps the \texttt{torch.nn.Linear} interface.

Second, we implement a batched conversion mechanism from quantized weights into their 16-bit form by assembling the weights into groups of 4 bits using the conversion techniques presented in \citet{kim2022sayselephantscantrun}. In the 4-bit regime, we can directly apply their 4-bit conversion intrinsics. However, to support batched conversion for all five bit-widths at no additional storage cost, we split the codes into five bit-plane buffers holding bits 1--2, 3, 4, 5--6 and 7--8, counting from the most significant bit. Serving $r$ bits therefore reads exactly $r$ bits per weight, and one stored model can switch bit-width at run time. Offline, the weights are permuted into the tensor-core register layout. This layout supports two deployment modes: loading only the first $r$ bits to serve an exact $r$-bit model with the memory footprint of a dedicated $r$-bit model, or keeping the full $c$-bit model resident and changing the bit-width dynamically during decoding, as suggested by \citet{chen2025progressivemixedprecisiondecodingefficient}.

Third, we use two kernels depending on the batch size. Since inference is memory-bandwidth bound for $b \le 4$, we use a direct-load kernel for streaming weights directly from memory while reading activations from L2, avoiding shared-memory staging and synchronization. For larger $b$, we use a batched kernel that reuses each dequantized weight fragment across $T \in \{8,16,32,64\}$ tokens, stages activations with double-buffered \texttt{cp.async}, and selects tile size and reduction split per layer shape, batch size, and bit-width via the CUDA occupancy API.

\section{Experiments}
\label{section:experiments}

\subsection{General Setup}
\label{sec:setup}

\textbf{Models \& Data.} 
We evaluate our method on the instruct-tuned versions of Llama3.1 8B~\citep{dubey2024llama3herdmodels} and Qwen3 8B~\citep{yang2025qwen3technicalreport}. 
To establish a point of comparison, we compare against independently quantized GPTQ, MatQuant, Any-Precision-LLM (AP), and AnyBCQ (AB) models.  Details about the hyperparameters in Appendix \ref{section:gptq-params}.

\textbf{Evaluation Datasets.} Following established methodology \citep{gptq}, we evaluate the models on commonsense reasoning on the following zero-shot tasks: ARC (ARC-c, ARC-e) \citep{arc_allenai}, HellaSwag \citep{zellers2019hellaswag}, PIQA \citep{DBLP:conf/aaai/piqa2020} and Winogrande \citep{sakaguchi2021winogrande}. Additionally, we evaluate generation quality on the non-chain-of-thought versions of GSM8K~\citep{gsm8k}, MMLU~\citep{hendrycks2020measuring}, and IFEval~\citep{zhou2023instructionfollowingevaluationlargelanguage}.

\textbf{Implementation Details.}
Since delivering an open-source version was a major goal, we built our code on top of the widely used PyTorch-based Hugging Face framework \citep{wolf2019huggingface} and integrated the serving infrastructure into vLLM~\citep{vllm}. All experiments were performed on an NVIDIA RTX A6000 (Ampere) using CUDA 12.4.

\subsection{Accuracy Evaluations for MatGPTQ}

We now examine the accuracy of \emph{MatGPTQ} across the models and benchmarks introduced previously. We compare MatGPTQ against independently quantized GPTQ, MatQuant, Any-Precision-LLM (AP) and AnyBCQ (AB). In contrast to MatQuant, which only quantizes FFN layers using OmniQuant, we quantize both FFN and attention layers using all methods.

For MatGPTQ and MatQuant, we use the target bit-width set $R = \{3,4,8\}$ with master bit-width $c = 8$, For MatGPTQ $\lambda_r$ is chosen to balance accuracy across bit-widths (see Appendix \ref{section:lambda-evaluation}) that chooses balance across bit-widths. For AP and AB, we start from a 3-bit seed model and progressively increase the bit-width up to 8 bits. We focus on this configuration because the 3–4 bit regime is particularly relevant for low-cost deployment, while still achieving strong recovery of model performance. Higher bit-widths nevertheless remain relevant, as full performance recovery is not guaranteed at 4 bits, especially on more challenging tasks \citep{kurtic2024give, park2024any, ICLR2026_ca9eaef0}. Moreover, higher bit-widths enable non-uniform inference settings in which different layers are assigned different precisions \citep{he2018amc, spdy, owl}. Finally, MatGPTQ uses asymmetric quantization to make the comparison more fair against the other methods, which are asymmetric.

A direct comparison against the original MatQuant framework is challenging, since neither source code nor models are publicly available. Furthermore, they omit evaluation hyperparameters, making reproduction substantially more difficult. We re-implemented the OmniQuant-based MatQuant approach to produce the MatQuant models, while AP and AB models were generated using their publicly available code.  Hyperparameters are detailed in Appendix~\ref{section:gptq-params}.

\begin{table*}[t]
\centering
\small
\setlength{\tabcolsep}{3.5pt}
\caption{Unified accuracy comparison of Llama3.1-8B-Instruct and Qwen3-8B across bitwidths, for MatGPTQ against dedicated GPTQ, MatQuant, Any-Precision-LLM (AP) and AnyBCQ (AB). CSR averages ARC-c, ARC-e, HellaSwag, PIQA and WinoGrande; IFEval averages its prompt- and instance-level strict accuracies. The mean over three seeds is reported. For each format, top methods within variance are marked in \textbf{bold}, and methods within variance of the full-precision model (full recovery) are \underline{underlined}.}
\label{tab:main}
\begin{tabular}{ll rrrrr rrrrr}
\toprule
 & & \multicolumn{5}{c}{Llama3.1-8B-Instruct} & \multicolumn{5}{c}{Qwen3-8B} \\
\cmidrule(lr){3-7}\cmidrule(lr){8-12}
Bits & Method & CSR & GSM8K & MMLU & IFEval & Avg & CSR & GSM8K & MMLU & IFEval & Avg \\
\midrule
16 & Baseline & 73.77 & 85.75 & 69.32 & 79.01 & 76.96 & 71.52 & 87.57 & 74.88 & 83.76 & 79.43 \\
\midrule
\multirow{6}{*}{8} & GPTQ & 73.80 & 84.84 & 69.32 & 78.53 & \underline{\textbf{76.62}} & 71.58 & 87.11 & 74.90 & 83.95 & \underline{\textbf{79.39}} \\
  & AP & 73.81 & 85.65 & 69.33 & 78.16 & \underline{\textbf{76.74}} & 71.57 & 87.43 & 74.90 & 84.48 & \underline{\textbf{79.60}} \\
  & AB & 73.60 & 84.10 & 69.11 & 78.07 & 76.22 & 71.74 & 87.49 & 74.71 & 84.48 & \underline{\textbf{79.60}} \\
  & MatQuant & 73.08 & 84.23 & 69.03 & 78.68 & \textbf{76.26} & 71.34 & 87.49 & 75.10 & 84.76 & \underline{\textbf{79.68}} \\
  & MatGPTQ & 73.75 & 84.69 & 69.07 & 77.98 & \textbf{76.37} & 71.30 & 87.36 & 74.73 & 84.65 & \underline{79.51} \\
\midrule
\multirow{6}{*}{4} & GPTQ & 73.66 & 81.27 & 68.65 & 76.79 & \textbf{75.09} & 71.45 & 86.96 & 73.81 & 83.71 & \textbf{78.98} \\
  & AP & 72.67 & 81.27 & 67.27 & 77.35 & 74.64 & 70.81 & 86.44 & 73.55 & 84.24 & \textbf{78.76} \\
  & AB & 72.84 & 80.49 & 67.55 & 77.48 & 74.59 & 70.84 & 86.48 & 73.49 & 84.62 & \textbf{78.86} \\
  & MatQuant & 71.70 & 78.97 & 66.05 & 77.12 & 73.46 & 70.28 & 86.60 & 73.49 & 83.64 & \textbf{78.50} \\
  & MatGPTQ & 73.08 & 82.84 & 68.33 & 77.23 & \textbf{75.37} & 71.28 & 87.14 & 73.64 & 84.09 & \underline{\textbf{79.04}} \\
  & \textcolor{gray}{GPTQ sliced} & \textcolor{gray}{72.49} & \textcolor{gray}{81.43} & \textcolor{gray}{66.78} & \textcolor{gray}{75.70} & \textcolor{gray}{74.10} & \textcolor{gray}{70.00} & \textcolor{gray}{84.61} & \textcolor{gray}{73.05} & \textcolor{gray}{85.99} & \textcolor{gray}{78.41} \\
\midrule
\multirow{6}{*}{3} & GPTQ & 71.72 & 69.90 & 63.62 & 70.82 & \textbf{69.02} & 69.32 & 80.74 & 70.66 & 79.23 & \textbf{74.99} \\
  & AP & 67.49 & 58.50 & 60.64 & 69.66 & 64.07 & 67.08 & 80.83 & 69.43 & 80.33 & \textbf{74.42} \\
  & AB & 71.29 & 64.39 & 63.90 & 75.73 & \textbf{68.83} & 69.50 & 76.35 & 69.11 & 77.62 & 73.14 \\
  & MatQuant & 55.02 & 42.58 & 48.19 & 52.80 & 49.65 & 67.19 & 77.04 & 66.96 & 77.86 & 72.26 \\
  & MatGPTQ & 70.45 & 68.28 & 62.96 & 73.62 & \textbf{68.83} & 68.41 & 80.46 & 69.08 & 79.43 & \textbf{74.35} \\
  & \textcolor{gray}{GPTQ sliced} & \textcolor{gray}{65.16} & \textcolor{gray}{33.51} & \textcolor{gray}{50.71} & \textcolor{gray}{55.25} & \textcolor{gray}{51.16} & \textcolor{gray}{57.89} & \textcolor{gray}{54.44} & \textcolor{gray}{57.63} & \textcolor{gray}{55.13} & \textcolor{gray}{56.27} \\
\midrule \midrule
\multirow{6}{*}{6} & GPTQ & 73.46 & 84.46 & 69.05 & 78.16 & \textbf{76.28} & 71.17 & 87.04 & 74.84 & 84.31 & \underline{\textbf{79.34}} \\
  & AP & 73.87 & 84.31 & 69.13 & 78.63 & \underline{\textbf{76.48}} & 71.55 & 87.45 & 74.97 & 84.41 & \underline{\textbf{79.59}} \\
  & AB & 73.63 & 84.26 & 69.07 & 77.94 & \textbf{76.23} & 71.65 & 87.29 & 74.77 & 83.77 & \underline{\textbf{79.37}} \\
  & MatQuant & 73.28 & 83.62 & 68.91 & 76.96 & 75.70 & 71.10 & 87.55 & 74.95 & 84.39 & \underline{\textbf{79.50}} \\
  & MatGPTQ & 73.69 & 84.82 & 68.93 & 77.77 & \textbf{76.30} & 71.46 & 87.34 & 74.76 & 83.60 & 79.29 \\
  & \textcolor{gray}{GPTQ sliced} & \textcolor{gray}{73.51} & \textcolor{gray}{84.53} & \textcolor{gray}{69.11} & \textcolor{gray}{78.78} & \textcolor{gray}{\underline{76.48}} & \textcolor{gray}{71.33} & \textcolor{gray}{87.72} & \textcolor{gray}{74.83} & \textcolor{gray}{84.31} & \textcolor{gray}{\underline{79.55}} \\
\bottomrule
\end{tabular}
\end{table*}

\textbf{Main Results.} In Table \ref{tab:main}, we see the downstream performance of the models. First, we see that MatGPTQ is either better than or within variance of the independently quantized GPTQ version for all optimized bitwidths. This generally ensures that quality is maintained even though the model is optimized for different bit-widths.
Second, we see in bold that MatGPTQ is within variance of the best-performing model for all bit-widths for both models, except for the 8-bit Llama3.1-8B-Instruct model, compared to MatQuant, AP, and AB.
Third, we see that MatGPTQ outperforms MatQuant at nearly every bit-width for both models, especially in the low-bit regime, where MatQuant models drop in performance and do not scale to quantizing Attention+FFNs.
Lastly, we apply our slicing approach to an independently quantized 8-bit GPTQ model to evaluate whether the multi-precision objective is necessary. As shown by the gray results, the sliced 8-bit GPTQ model maintains performance at 6 and 8 bits, but increasingly degrades at 4 bits and drops substantially at 3 bits. 
Therefore, the MatGPTQ objective is necessary to achieve strong performance across bit-widths, while remaining competitive with AB and AP using the simpler and improved MatQuant representation and a smaller overall checkpoint footprint (see Table~\ref{tab:method_comparison}).

\textbf{Interpolation.} As mentioned in~\citet{matryoshka_quantization}, also MatGPTQ shows strong performance when interpolating bit-widths that are not directly optimized for. As we see in Table \ref{tab:main} both models are on par with the GPTQ baseline. Since we provide CUDA kernel support for 2, 3, 4, 6, and 8 bits, we are also interested in being able to use a 6-bit model. We can see in Table~\ref{tab:main} that the interpolated 6-bit MatGPTQ model remains within variance of the independently quantized GPTQ model and achieves full performance recovery for both models, showcasing strong interpolation performance.

\subsection{CUDA Kernel Performance}

\begin{figure}[h]
    \centering
    \includegraphics[width=1\linewidth]{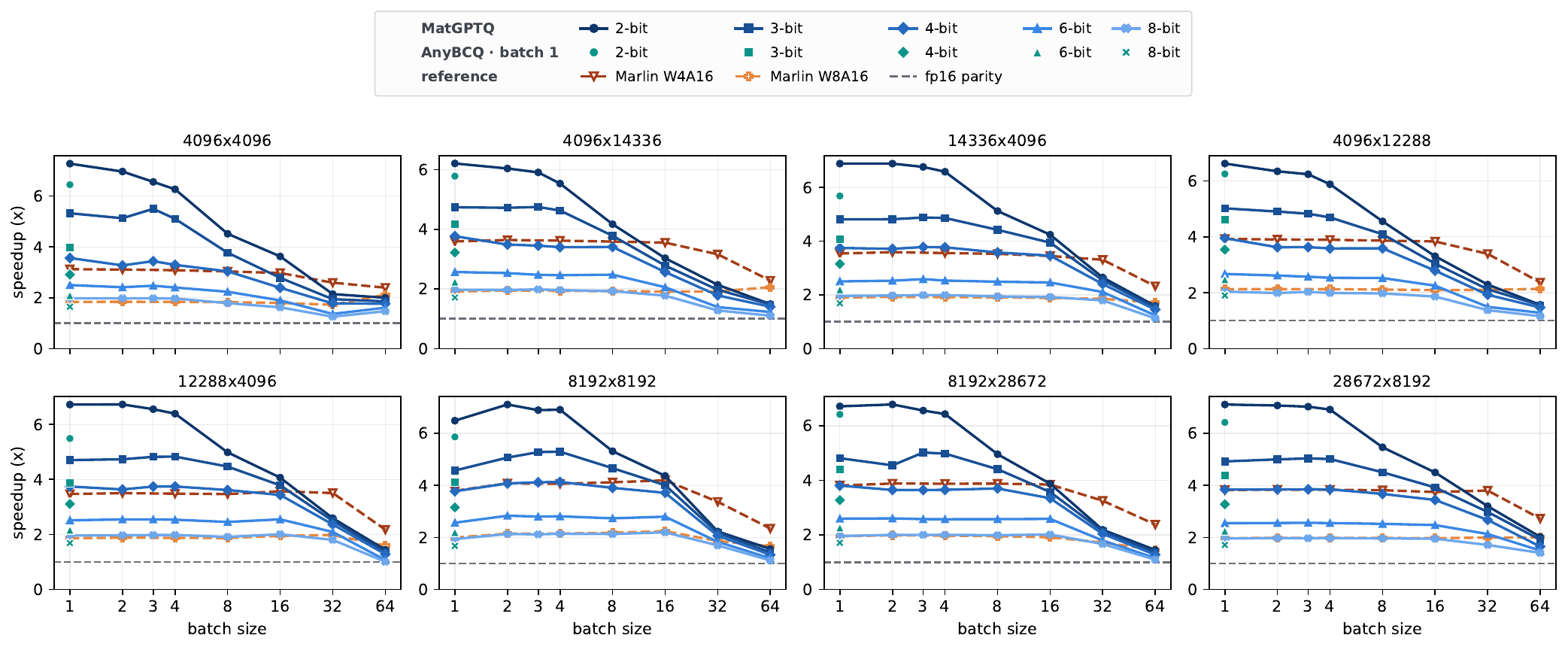}
    \caption{The figure shows the kernel speedup over FP16 \texttt{torch.matmul} (cuBLAS). The kernels are evaluated at 2--8 bit precision for several representative matrix shapes across different batch sizes. AnyBCQ and Marlin W4A16/W8A16 are included as reference baselines. However, AnyBCQ results are only shown for batch size 1.}
    \label{fig:kernel-perf}
\end{figure}

We benchmark the kernel latency for common weight matrix sizes of models such as Llama3.1-8B, Qwen3-8B, and Llama3-70B, while varying the input size. These latency measurements were taken in an isolated environment, purely evaluating kernel performance and ignoring overhead introduced by Hugging Face or other libraries.
We use \texttt{torch.matmul} (cuBLAS) as a baseline for comparison. All configurations use a group size of 128. As seen in Figure \ref{fig:kernel-perf}, our kernels achieve speedups between 2x and 6.5x in the memory-bound regime. The peak speedup is achieved for 2-bit single-batch multiplication, owing to the memory-bound nature of this setting. At higher batch sizes, the kernel transitions to a compute-bound regime, and we observe slowdowns at around batch size 32.

Furthermore, we integrated the AnyBCQ kernels into our benchmark harness, as AnyBCQ reports faster kernels than Any-Precision-LLM, as another point of comparison \citep{ICLR2026_ca9eaef0}. As seen in Figure \ref{fig:kernel-perf}, at batch size 1, AnyBCQ kernels, visualized with green dots, are consistently slower than MatGPTQ kernels for all bit-widths across all shapes. 
At larger batch sizes, AnyBCQ provides no dedicated kernel and falls back to dequantization followed by cuBLAS, whereas our kernels continue to operate directly on the packed representation.
This leads to significant performance degradation, highlighting our contribution.

Additionally, we benchmarked the highly optimized vLLM-Marlin kernel for 4- and 8-bit quantization. As shown in Figure~\ref{fig:kernel-perf}, indicated by the red and orange lines, MatGPTQ at 4 and 8 bits remains competitive with the dedicated Marlin kernels and only falls behind from batch size 16 onward.

\subsection{End-to-End Performance}

\begin{figure}[t]
  \centering

  \begin{minipage}[t]{0.49\linewidth}
    \centering
    \includegraphics[
      width=\linewidth,
      height=4.2cm,
      keepaspectratio
    ]{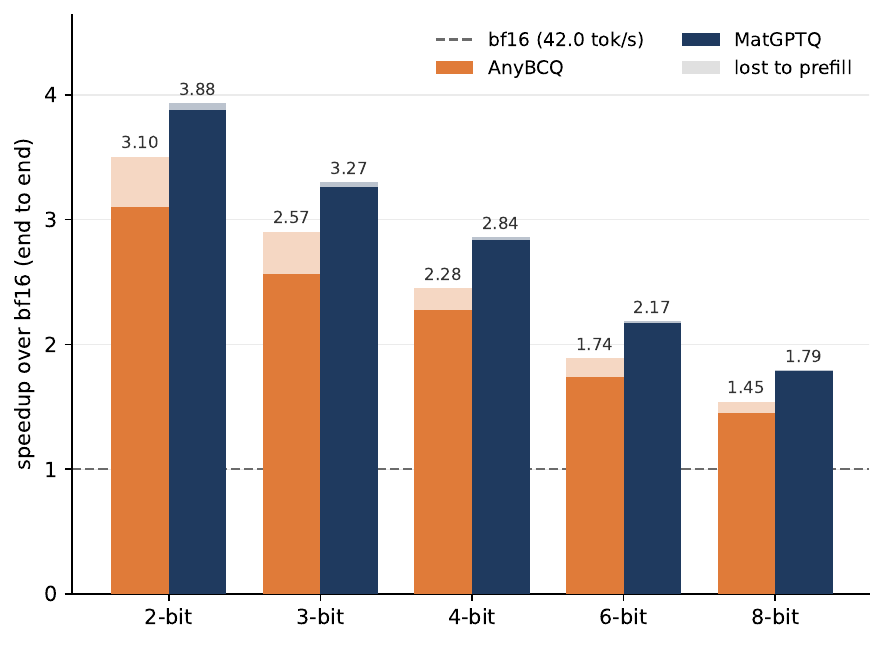}
    \captionof{figure}{End-to-end speedups: Transformers -- Llama3.1-8B-Instruct, Prompt/Decode 32/128.}
    \label{fig:e2e_short}
  \end{minipage}
  \hfill
  \begin{minipage}[t]{0.49\linewidth}
    \centering
    \includegraphics[
      width=\linewidth,
      height=4.2cm,
      keepaspectratio
    ]{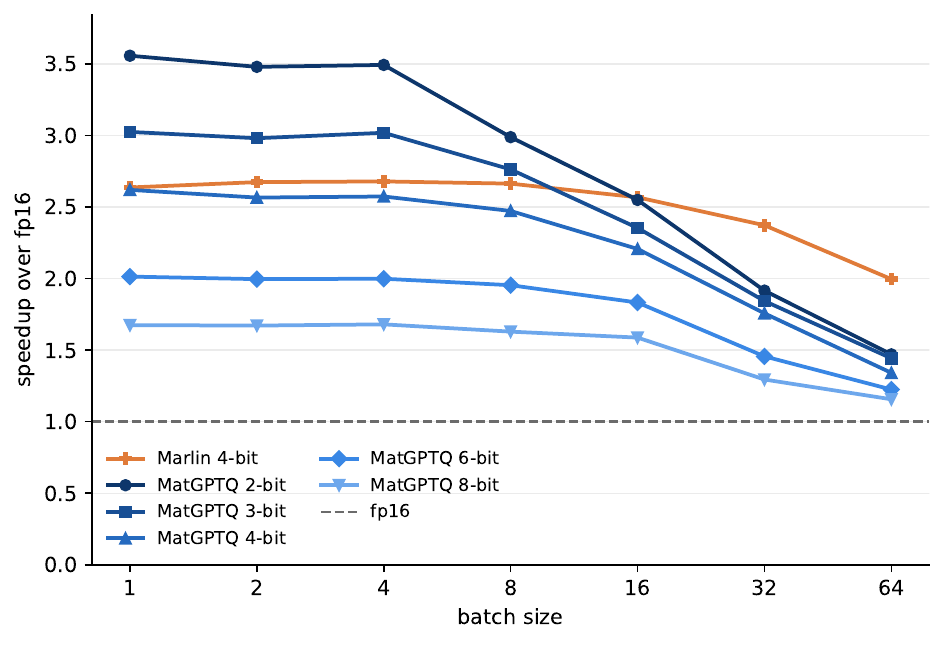}
    \captionof{figure}{End-to-end speedups: vLLM -- Llama3.1-8B-Instruct, Prompt/Decode 32/128.}
    \label{fig:e2e_vllm}
  \end{minipage}

\end{figure}

We integrate our kernels into the widely used PyTorch-based Hugging Face framework~\citep{wolf2019huggingface}, as well as vLLM~\citep{vllm} for real-world deployment. Using those two system adaptations, we measure end-to-end throughput (token/s) while varying the model bit-width. For the Hugging Face framework, we utilize \texttt{torch.compile} under \texttt{reduce-overhead} and \texttt{full-graph} settings, while for vLLM we utilize the default torch.compile mode.

In Figure~\ref{fig:e2e_short}, we can see that the speed-up over the 16-bit version naturally diminishes as bit-widths increase. In the low-bit regime (2--4 bits), our kernels achieve end-to-end speedups of $\approx$2.8x to 3.9x. Furthermore, we see that we outperform the AnyBCQ kernels at every bit-width and can clearly see the difference in prefill time, especially in the low-bit regime. Additional results for larger prompt/decoding example can be found in Appendix \ref{sec:kernel_details}.

These improvements also translate to our vLLM integration. Overall speedups between 1.7x and 3.5x can be achieved in the memory-bound regime. Furthermore, we include results using the highly optimized vLLM-Marlin kernel execution at 4-bit precision. As shown in Figure~\ref{fig:e2e_vllm}, MatGPTQ remains competitive at smaller batch sizes and falls behind when leaving the memory-bound regime.

\section{Discussion \& Limitations}
In this work, we close the gaps in nested LLM inference by introducing MatGPTQ, the first practical and efficient framework for post-training Matryoshka quantization. By adapting the GPTQ algorithm to optimize a single model jointly for multiple target precisions, MatGPTQ enables ``sliceable'' quantization in a one-shot manner. Furthermore, the improved representation enables true-prefix slicing. Together with a kernel that supports both prefill and decoding at any batch size, we address the limitations of current state-of-the-art methods. Therefore, a single quantized checkpoint can be obtained quickly and adapted at runtime to meet diverse memory and latency budgets across various hardware without the overhead of storing and maintaining multiple models.
Our extensive evaluation demonstrates that this flexibility comes at virtually no cost to accuracy relative to the GPTQ baseline, outperforming or being on par with current state-of-the-art methods, while using the simplest format out of all methods.

In addition, we provide the first open-source implementation of Matryoshka quantization, built on the Hugging Face framework, along with efficient CUDA kernels and a vLLM integration.

\textbf{Limitations and Future Work.} Our experiments focused on a limited range of bit-widths and on integer quantization. Achieving high accuracy in the ultra-low-bit regime remains challenging with the current GPTQ algorithm at standard group sizes. In future work, we will explore integrating Matryoshka objectives with quantization methods better suited for ultra-low precisions, as well as floating-point quantization. Furthermore, while our CUDA kernels currently target NVIDIA Ampere, Ada and Hopper GPUs, future work will expand support to newer architectures such as Blackwell.

\bibliography{iclr2027_conference}
\bibliographystyle{iclr2027_conference}

\newpage

\appendix

\section{MatGPTQ implementation details}

This section introduces additional implementation details about the MatGPTQ algorithm including the improved representation, the MSE-based scale search and the generation of quantized weights.

\subsection{Improved nested representation}
\label{sec:improved_slicing}

\begin{figure}[h]
  \centering
  \begin{subfigure}[b]{0.595\linewidth}
    \centering
    \includegraphics[width=\linewidth]{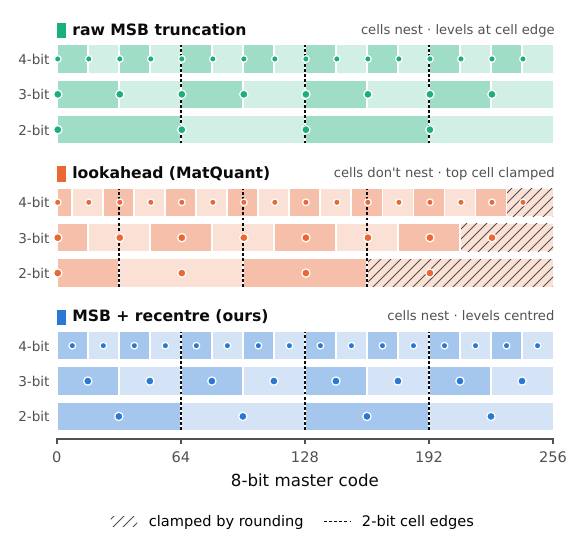}
    \caption{Which master codes share a level.}
    \label{fig:slicing_mapping}
  \end{subfigure}\hfill
  \begin{subfigure}[b]{0.395\linewidth}
    \centering
    \includegraphics[width=\linewidth]{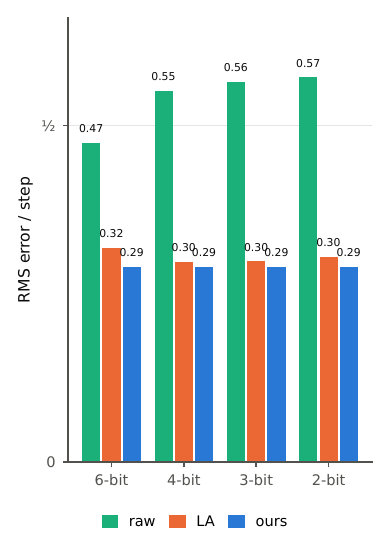}
    \caption{RMS slice error.}
    \label{fig:slicing_rms}
  \end{subfigure}
    \caption{
    Slicing an 8-bit master code to lower bit-widths.
    \subref{fig:slicing_mapping}~Raw truncation, MatQuant rounding, and our recentered truncation. Shaded regions indicate the corresponding cells, while dots denote the reconstructed quantized values. Our method preserves nested cells while centering the reconstruction within each cell.
    \subref{fig:slicing_rms}~RMS reconstruction error across bit-widths. Our recentered truncation achieves the lowest error.
    }
  \label{fig:slicing}
\end{figure}

The general idea behind slicing in MatQuant is to approximate a $c$-bit model using $r$ bits. First, a right shift keeps the $r$ most significant bits, effectively slicing the $c$-bit code. A subsequent left shift maps this $r$-bit prefix back into the $c$-bit code space, yielding the reconstructed quantized value. However, raw MSB truncation systematically reconstructs values at the lower edge of each cell, as seen in Figure~\ref{fig:slicing}, indicated by the green grid. As a result, the reconstructed $r$-bit values are, on average, farther from the corresponding $c$-bit values, leading to a higher root-mean-square (RMS) reconstruction error.

To avoid this issue, the MatQuant authors push values into higher buckets~\citep{matryoshka_quantization}, which introduces the following slicing algorithm: MatQuant slicing (Equation~\ref{eq:slicing}) extracts the $r$ most significant bits, uses the $(r+1)^{\mathrm{th}}$ bit for rounding, and then left-shifts the result back into the $c$-bit code space to obtain the reconstructed quantized value. Since the rounding step can lead to overflow, $\mathrm{clamp}(\cdot)$ is needed to ensure that the sliced values remain within the valid $r$-bit range. This can be seen in Figure~\ref{fig:slicing}, indicated by the orange grid.

To avoid extracting an additional bit and relying on the $\mathrm{clamp}(\cdot)$ function, we analyzed raw MSB truncation. Since the right and left shifts reconstruct the sliced value at the lower edge of each cell, we shift this reconstructed quantized value toward the center of the cell. This can be achieved by simply adding $\delta_r$, resulting in the blue grid shown in Figure~\ref{fig:slicing}. By centering the reconstructed value within each cell, the distances to the corresponding $c$-bit values are balanced, which minimizes the RMS reconstruction error for the fixed nested partition. Additionally, $\delta_r$ can be absorbed into the zero-point and therefore naturally integrated without additional overhead.

\subsection{MatGPTQ adaptation}
\label{section:perf-alg2}

This section explains the adaptations of the GPTQ algorithm. Algorithm~\ref{alg:scale-search} adapts GPTQ's scale search to the multi-precision setting using an MSE-based objective that incorporates the bit-width-specific weights $\lambda_r$.

\begin{algorithm}[ht!]
  \caption{$\mathrm{fit}$: shared $(\mathbf{s},\mathbf{z})$ of a weight group, chosen against every served bit-width $r\in R$ with weights $\lambda_r$ (per row)}
  \label{alg:scale-search}
  \begin{algorithmic}
    \STATE \textbf{input} group $\mathbf{X}\in\mathbb{R}^{d_{\mathrm{row}}\times g}$, Hessian diagonal $\mathbf{h}$, bit-widths $R$, weights $\lambda_R$, grid $G$, offsets $K$, shrink $\sigma$
    \STATE $\mathbf{x}_{\min},\mathbf{x}_{\max}\gets$ row-wise min, max of $\mathbf{X}$;\quad $d\gets c-\min R$
    \STATE $\mathcal{S}\gets\bigl\{\,p\,(\mathbf{x}_{\max}-\mathbf{x}_{\min})/(2^{c}-1)\;:\;p\in[\sigma,1],\ G \text{ steps}\,\bigr\}$ \hfill $\triangleright$ shrunk scales
    \STATE $\mathcal{Z}(\mathbf{s})\gets\bigl\{\,\mathrm{round}(-\mathbf{x}_{\min}/\mathbf{s})+k\,2^{d}/K\;:\;k<K\,\bigr\}$ \hfill $\triangleright$ shifted zeros
    \FOR{$\mathbf{s}\in\mathcal{S}$, $\mathbf{z}\in\mathcal{Z}(\mathbf{s})$}
      \STATE $\mathbf{Q}\gets\mathrm{clip}\bigl(\mathrm{round}(\mathbf{X}/\mathbf{s}+\mathbf{z}),\,0,\,2^{c}-1\bigr)$ \hfill $\triangleright$ round-to-nearest proxy
      \STATE $\mathcal{E}(\mathbf{s},\mathbf{z})\gets\sum_{r\in R}\lambda_r\,\bigl\|\sqrt{\mathbf{h}}\odot\bigl(\mathbf{X}-\mathbf{s}(S'(\mathbf{Q},r)+\delta_r-\mathbf{z)}\bigr)\bigr\|_2^{2}$ \hfill $\triangleright$ weighted error of all slices
    \ENDFOR
    \STATE \textbf{return} $\arg\min_{\mathbf{s},\mathbf{z}}\mathcal{E}(\mathbf{s},\mathbf{z})$
  \end{algorithmic}
\end{algorithm}

Algorithm \ref{alg:quant-col} returns the quantized weights, by first generating all possible quantized values in the range $[0, Q_{\max}]$ for each individual weight. For every target bit-width, these candidates are then dequantized using the respective scale and zero-point. The reconstruction error is computed per weight and accumulated across all candidates, weighted by their respective importance. This results in the total error for each potential quantized value of every weight. Finally, for each weight, the quantized value with the lowest total error is selected. Even though this is a brute-force approach and therefore memory and runtime concerns exist, the overhead is generally minimal.

\begin{algorithm}[ht!]
  \caption{$\text{quant}_{\text{MatGPTQ}}$: one master code per weight, chosen for every served bit-width at once. $S'(q,r)$ is the $r$-bit reconstruction of code $q$.}
  \label{alg:quant-col}
  \begin{algorithmic}
    \STATE \textbf{input} column $\mathbf{w}\in\mathbb{R}^{d_{\mathrm{row}}}$, scale $s$, zero-point $z$, bit-widths $R$, weights $\lambda_R$, master bit-width $c$
    \STATE $\displaystyle \mathbf{q}\gets \arg\min_{0\le q<2^{c}} \sum_{r\in R}\lambda_r\bigl(\mathbf{w}-s(S'(q,r) + \delta_r -z)\bigr)^{2}$ \hfill $\triangleright$ elementwise over rows
    \STATE \textbf{return} $\mathbf{q}$
  \end{algorithmic}
\end{algorithm}

\paragraph{Runtime \& memory considerations for \texttt{quant}$_{MatGPTQ}(\cdot)$.} With Algorithm \ref{alg:quant-col} we want to optimize the quantization grid for multiple bit-widths. We decided to take a simple brute-force approach instead of relying on any heuristics, as this approach is still highly vectorizable and the resulting overhead appeared acceptable considering our target of optimizing for multiple bit-widths.

\begin{table}[h]
    \centering
    \scriptsize
    \caption{Runtime comparison of per-column functions and full quantization passes. GPTQ$\times$3 is the sum of separate 3-, 4-, and 8-bit GPTQ passes. Speedup is (GPTQ$\times$3)/MatGPTQ.}    \begin{tabular}{l|rr|rrrr}
    \toprule
    & \multicolumn{2}{c|}{Per-column function (ms)} & \multicolumn{4}{c}{Complete pass (s)} \\ 
    \cmidrule{2-3} \cmidrule{4-7}
    Shape & \texttt{quant} & \texttt{quant\_MatGPTQ} & GPTQ & GPTQ$\times$3 & MatGPTQ & Speedup \\ 
    \midrule
    k/v (1024$\times$4096) & 0.0599 & 0.5360 & 1.45 & 4.34 & 3.59 & 1.21$\times$ \\
    q/o (4096$\times$4096) & 0.0601 & 0.5368 & 1.47 & 4.41 & 3.59 & 1.23$\times$ \\
    gate/up (14336$\times$4096) & 0.0610 & 1.3444 & 1.41 & 4.24 & 6.58 & 0.64$\times$ \\
    down (4096$\times$14336) & 0.0601 & 0.5368 & 5.41 & 16.24 & 12.45 & 1.30$\times$ \\
    \midrule
    Layer total & -- & -- & 14.07 & 42.21 & 39.96 & 1.06$\times$ \\
    \bottomrule
    \end{tabular}
    \label{tab:matgptq-complete-pass-runtime}
\end{table}

To validate this, we measured execution times for layer sizes of a Llama3.1-8B-Instruct model, both for the isolated per-column functions and complete (Mat)GPTQ passes. Generally, we can see that \texttt{quant\_MatGPTQ} is slower than the native \texttt{quant} function. Therefore, individual MatGPTQ quantization passes are slower than GPTQ passes. However, when comparing the cost of quantizing three separate GPTQ models with one MatGPTQ model, we see that MatGPTQ is faster on average.

Furthermore, Algorithm \ref{alg:quant-col} would indeed have higher peak memory usage than basic rounding in GPTQ, due to maintaining a matrix size of $2^c \times d_{row} \times d_{col}$. However, in practice, both the original GPTQ and our MatGPTQ implementation quantize only one column at a time. Therefore, the matrix has size $2^c \times d_{row} \times 1$, which, for a matrix of $4096\times 4096$, results in only 4 MB of overhead when stored in FP32 ($256 \times 4096 \times 1 \times 4$ bytes). This overhead is therefore insignificant in practice.

\section{Hyperparameter details}
This section presents the hyperparameters used for the respective methods.
\label{section:gptq-params}

\textbf{MatGPTQ and GPTQ.} For both the native GPTQ and MatGPTQ, we use the Fineweb-Edu \citep{penedo2024finewebdatasetsdecantingweb} dataset as calibration data. Specifically, we use 2,097,152 tokens, corresponding to 1024 samples with a sequence length of 2048. Furthermore, we use a group size of 128, a dampening factor of 0.01, MSE-based scale search and asymmetric quantization. For completeness, we optimized for bit-widths $R = \{3, 4, 8\}$ and used the most balanced config as elaborated in Section \ref{section:lambda-evaluation}

\textbf{MatQuant.} For MatQuant~\citep{matryoshka_quantization}, we followed the specific hyperparameters provided by the authors: For Llama3.1-8B-Instruct and Qwen3-8B we also used $\lambda_3 = 1$, $\lambda_4 = 0.1$, and $\lambda_8 = 0.1$. We used their constant learning rate of $10^{-3}$ and a batch size of 4. We only differ by using asymmetric quantization and optimizing for $r = \{3, 4, 8\}$. Additionally, we apply grouping of size 128 to make the comparison even fairer, even though they never mention anything regarding grouping.

\textbf{Any-Precision-LLM.} For Any-Precision-LLM~\citep{park2024any}, we use the authors' reference implementation and its default hyperparameters. The sensitivity gradients and the k-means seed model are computed on the Fineweb-Edu \citep{penedo2024finewebdatasetsdecantingweb} dataset with 1024 calibration samples of sequence length 2048. We use a seed precision of 3 and a parent precision of 8, so that a single checkpoint supports $r \in \{3, \dots, 8\}$.

\textbf{AnyBCQ.} For AnyBCQ~\citep{ICLR2026_ca9eaef0}, we use the authors' reference implementation and its default hyperparameters. The binary weight planes and their per-group scales are fitted on the Fineweb-Edu \citep{penedo2024finewebdatasetsdecantingweb} dataset with 1024 calibration samples of sequence length 2048, using a group size of 128 and asymmetric quantization, with an input probability of 1.0 for Llama-3.1-8B-Inst and 0.5 for Qwen3-8B. We quantize from 3 to 8 bits, so that a single checkpoint stores 8 binary planes together with per-bit-width scales and offsets and supports $r \in \{3, \dots, 8\}$, all other settings are left at their defaults.

\section{Detailed Accuracy Results}
\label{sec:variance}

Tables \ref{tab:full_results_asym_llama_3_1_8b_it} and \ref{tab:full_results_asym_qwen3_8b} present detailed evaluation results of Llama3.1-8B-Instruct and Qwen3-8B with MatGPTQ versus independently quantized GPTQ, MatQuant, Any-Precision-LLM (AP) and AnyBCQ models. All methods use asymmetric schemes.

\begin{table*}
\caption{Evaluation results for asymmetric MatGPTQ, GPTQ, MatQuant, Any-Precision-LLM (AP) and AnyBCQ (AB) on Llama3.1-8B-Instruct. Accuracy on the evaluation tasks ($\uparrow$): ARC-c (ARC-Challenge), ARC-e (ARC-Easy), HellaSwag, PIQA, WinoG (WinoGrande), GSM8K, MMLU, and IF-p and IF-i, the prompt- and instance-level strict IFEval accuracies. The mean over three calibration seeds is reported.}
\centering
\setlength{\tabcolsep}{3.5pt}
\tiny
\begin{tabular}{cl*{9}{c}}
\toprule
\multicolumn{11}{c}{Llama3.1-8B-Instruct} \\
\midrule
Bitwidth & Method & ARC-c & ARC-e & HellaSwag & PIQA & WinoG & GSM8K & MMLU & IF-p & IF-i \\
\midrule
16 & & 55.20 & 79.55 & 79.22 & 81.01 & 73.88 & 85.75 & 69.32 & 75.42 & 82.61 \\ \midrule
\multirow{6}{*}{8} & GPTQ & $54.95_{0.19}$ & $79.80_{0.13}$ & $79.21_{0.03}$ & $80.85_{0.05}$ & $74.19_{0.28}$ & $84.84_{0.50}$ & $69.32_{0.05}$ & $75.05_{0.71}$ & $82.01_{0.61}$ \\
 & AP & $55.15_{0.27}$ & $79.67_{0.11}$ & $79.22_{0.04}$ & $80.96_{0.14}$ & $74.06_{0.05}$ & $85.65_{0.42}$ & $69.33_{0.06}$ & $74.43_{0.59}$ & $81.89_{0.32}$ \\
 & AB & $55.83_{0.74}$ & $79.66_{0.53}$ & $78.39_{0.15}$ & $81.16_{0.03}$ & $72.98_{0.12}$ & $84.10_{0.66}$ & $69.11_{0.15}$ & $74.49_{0.67}$ & $81.65_{0.62}$ \\
 & MatQuant & $53.90_{0.13}$ & $78.63_{0.11}$ & $78.93_{0.02}$ & $81.08_{0.03}$ & $72.88_{0.24}$ & $84.23_{0.08}$ & $69.03_{0.04}$ & $75.11_{0.11}$ & $82.25_{0.24}$ \\
 & MatGPTQ & $54.93_{0.27}$ & $79.83_{0.19}$ & $78.87_{0.06}$ & $81.02_{0.17}$ & $74.11_{0.06}$ & $84.69_{0.23}$ & $69.07_{0.15}$ & $74.23_{0.43}$ & $81.73_{0.51}$ \\ \midrule
\multirow{6}{*}{4} & GPTQ & $55.03_{0.34}$ & $79.92_{0.41}$ & $78.27_{0.14}$ & $80.79_{0.27}$ & $74.27_{0.40}$ & $81.27_{1.02}$ & $68.65_{0.18}$ & $73.01_{1.74}$ & $80.58_{1.23}$ \\
 & AP & $53.95_{0.51}$ & $78.56_{0.49}$ & $77.83_{0.26}$ & $80.45_{0.49}$ & $72.53_{1.32}$ & $81.27_{1.07}$ & $67.27_{0.32}$ & $73.69_{1.20}$ & $81.02_{0.84}$ \\
 & AB & $54.12_{0.58}$ & $79.24_{0.38}$ & $77.51_{0.10}$ & $80.65_{0.08}$ & $72.69_{0.34}$ & $80.49_{0.16}$ & $67.55_{0.08}$ & $73.75_{0.37}$ & $81.22_{0.30}$ \\
 & MatQuant & $51.59_{0.18}$ & $78.03_{0.17}$ & $77.31_{0.08}$ & $79.83_{0.08}$ & $71.74_{0.44}$ & $78.97_{0.29}$ & $66.05_{0.10}$ & $73.26_{0.59}$ & $80.98_{0.50}$ \\
 & MatGPTQ & $53.96_{0.50}$ & $78.65_{0.78}$ & $78.42_{0.10}$ & $80.76_{0.11}$ & $73.62_{0.72}$ & $82.84_{0.94}$ & $68.33_{0.27}$ & $73.42_{1.40}$ & $81.03_{0.99}$ \\
 & \textcolor{gray}{GPTQ sliced} & \textcolor{gray}{$52.73_{0.30}$} & \textcolor{gray}{$78.11_{0.41}$} & \textcolor{gray}{$78.17_{0.12}$} & \textcolor{gray}{$79.82_{0.23}$} & \textcolor{gray}{$73.64_{0.43}$} & \textcolor{gray}{$81.43_{1.01}$} & \textcolor{gray}{$66.78_{0.20}$} & \textcolor{gray}{$71.53_{1.85}$} & \textcolor{gray}{$79.86_{1.20}$} \\ \midrule
\multirow{6}{*}{3} & GPTQ & $53.58_{1.37}$ & $78.32_{1.25}$ & $76.21_{0.25}$ & $78.56_{0.51}$ & $71.90_{0.72}$ & $69.90_{0.84}$ & $63.62_{0.35}$ & $65.99_{1.55}$ & $75.66_{1.27}$ \\
 & AP & $47.38_{2.69}$ & $72.57_{2.24}$ & $73.24_{1.62}$ & $76.57_{1.56}$ & $67.69_{1.28}$ & $58.50_{3.11}$ & $60.64_{0.63}$ & $64.94_{0.91}$ & $74.38_{0.82}$ \\
 & AB & $51.68_{0.18}$ & $77.69_{0.11}$ & $75.23_{0.07}$ & $79.45_{0.17}$ & $72.40_{0.12}$ & $64.39_{0.36}$ & $63.90_{0.08}$ & $71.72_{0.64}$ & $79.74_{0.52}$ \\
 & MatQuant & $29.81_{0.55}$ & $49.09_{0.98}$ & $67.94_{0.23}$ & $65.14_{0.22}$ & $63.14_{0.36}$ & $42.58_{0.94}$ & $48.19_{0.08}$ & $47.01_{1.87}$ & $58.59_{1.63}$ \\
 & MatGPTQ & $50.24_{0.38}$ & $75.88_{0.97}$ & $75.26_{0.28}$ & $78.82_{0.38}$ & $72.04_{0.65}$ & $68.28_{0.75}$ & $62.96_{0.55}$ & $69.54_{2.67}$ & $77.70_{2.11}$ \\
 & \textcolor{gray}{GPTQ sliced} & \textcolor{gray}{$43.94_{1.32}$} & \textcolor{gray}{$66.04_{1.21}$} & \textcolor{gray}{$71.34_{0.23}$} & \textcolor{gray}{$76.44_{0.53}$} & \textcolor{gray}{$68.03_{0.71}$} & \textcolor{gray}{$33.51_{0.84}$} & \textcolor{gray}{$50.71_{0.40}$} & \textcolor{gray}{$49.35_{1.50}$} & \textcolor{gray}{$61.15_{1.24}$} \\
\midrule \midrule
\multirow{6}{*}{6} & GPTQ & $54.18_{0.35}$ & $79.12_{0.44}$ & $79.23_{0.01}$ & $81.07_{0.15}$ & $73.72_{0.30}$ & $84.46_{0.40}$ & $69.05_{0.16}$ & $74.31_{0.55}$ & $82.01_{0.40}$ \\
 & AP & $55.32_{0.44}$ & $79.46_{0.15}$ & $79.24_{0.11}$ & $81.08_{0.30}$ & $74.24_{0.55}$ & $84.31_{0.42}$ & $69.13_{0.16}$ & $75.05_{0.81}$ & $82.21_{0.88}$ \\
 & AB & $55.49_{0.13}$ & $80.15_{0.38}$ & $78.48_{0.14}$ & $80.90_{0.05}$ & $73.14_{0.58}$ & $84.26_{0.77}$ & $69.07_{0.13}$ & $74.31_{1.95}$ & $81.57_{1.14}$ \\
 & MatQuant & $54.41_{0.05}$ & $78.37_{0.04}$ & $78.89_{0.01}$ & $81.34_{0.11}$ & $73.40_{0.16}$ & $83.62_{0.15}$ & $68.91_{0.04}$ & $73.07_{0.87}$ & $80.86_{0.87}$ \\
 & MatGPTQ & $54.88_{0.42}$ & $79.60_{0.18}$ & $78.82_{0.07}$ & $81.01_{0.18}$ & $74.14_{0.49}$ & $84.82_{1.06}$ & $68.93_{0.12}$ & $73.94_{0.61}$ & $81.61_{0.60}$ \\
 & \textcolor{gray}{GPTQ sliced} & \textcolor{gray}{$54.52_{0.33}$} & \textcolor{gray}{$79.55_{0.45}$} & \textcolor{gray}{$79.18_{0.01}$} & \textcolor{gray}{$81.23_{0.17}$} & \textcolor{gray}{$73.09_{0.31}$} & \textcolor{gray}{$84.53_{0.38}$} & \textcolor{gray}{$69.11_{0.10}$} & \textcolor{gray}{$75.42_{0.58}$} & \textcolor{gray}{$82.13_{0.41}$} \\
\bottomrule
\end{tabular}
\label{tab:full_results_asym_llama_3_1_8b_it}
\end{table*}

\begin{table*}
\caption{Evaluation results for asymmetric MatGPTQ, GPTQ, MatQuant, Any-Precision-LLM (AP) and AnyBCQ (AB) on Qwen3-8B. Accuracy on the evaluation tasks ($\uparrow$): ARC-c (ARC-Challenge), ARC-e (ARC-Easy), HellaSwag, PIQA, WinoG (WinoGrande), GSM8K, MMLU, and IF-p and IF-i, the prompt- and instance-level strict IFEval accuracies. The mean over three calibration seeds is reported.}
\centering
\setlength{\tabcolsep}{3.5pt}
\tiny
\begin{tabular}{cl*{9}{c}}
\toprule
\multicolumn{11}{c}{Qwen3-8B} \\
\midrule
Bitwidth & Method & ARC-c & ARC-e & HellaSwag & PIQA & WinoG & GSM8K & MMLU & IF-p & IF-i \\
\midrule
16 & & 56.48 & 80.64 & 74.93 & 77.69 & 67.88 & 87.57 & 74.88 & 80.59 & 86.93 \\ \midrule
\multirow{6}{*}{8} & GPTQ & $56.66_{0.27}$ & $80.81_{0.07}$ & $74.98_{0.04}$ & $77.58_{0.06}$ & $67.88_{0.10}$ & $87.11_{0.16}$ & $74.90_{0.04}$ & $80.96_{0.59}$ & $86.93_{0.38}$ \\
 & AP & $56.52_{0.11}$ & $80.92_{0.09}$ & $74.93_{0.03}$ & $77.70_{0.05}$ & $67.80_{0.37}$ & $87.43_{0.28}$ & $74.90_{0.05}$ & $81.55_{0.36}$ & $87.41_{0.35}$ \\
 & AB & $56.40_{0.78}$ & $80.88_{0.57}$ & $74.72_{0.24}$ & $77.51_{0.11}$ & $69.17_{0.41}$ & $87.49_{1.14}$ & $74.71_{0.25}$ & $81.52_{1.03}$ & $87.45_{0.50}$ \\
 & MatQuant & $56.26_{0.29}$ & $80.56_{0.07}$ & $74.75_{0.08}$ & $77.27_{0.12}$ & $67.86_{0.31}$ & $87.49_{0.33}$ & $75.10_{0.07}$ & $82.07_{0.23}$ & $87.46_{0.14}$ \\
 & MatGPTQ & $56.03_{0.65}$ & $80.13_{0.28}$ & $74.68_{0.03}$ & $77.53_{0.20}$ & $68.14_{0.39}$ & $87.36_{0.23}$ & $74.73_{0.12}$ & $81.76_{0.28}$ & $87.53_{0.42}$ \\ \midrule
\multirow{6}{*}{4} & GPTQ & $55.89_{0.55}$ & $80.60_{0.90}$ & $74.66_{0.13}$ & $77.53_{0.40}$ & $68.59_{0.49}$ & $86.96_{0.39}$ & $73.81_{0.23}$ & $80.96_{0.83}$ & $86.45_{0.57}$ \\
 & AP & $55.34_{0.42}$ & $79.32_{0.68}$ & $73.77_{0.23}$ & $77.38_{0.38}$ & $68.26_{0.66}$ & $86.44_{0.76}$ & $73.55_{0.29}$ & $81.40_{0.90}$ & $87.07_{0.52}$ \\
 & AB & $55.20_{0.44}$ & $79.63_{0.29}$ & $73.55_{0.12}$ & $77.86_{0.22}$ & $67.96_{0.44}$ & $86.48_{0.04}$ & $73.49_{0.02}$ & $81.95_{0.75}$ & $87.29_{0.52}$ \\
 & MatQuant & $53.46_{0.42}$ & $79.96_{0.19}$ & $73.32_{0.14}$ & $77.16_{0.30}$ & $67.48_{0.60}$ & $86.60_{0.47}$ & $73.49_{0.16}$ & $80.70_{0.98}$ & $86.57_{0.63}$ \\
 & MatGPTQ & $55.97_{0.62}$ & $80.18_{0.45}$ & $73.93_{0.05}$ & $77.69_{0.05}$ & $68.64_{1.07}$ & $87.14_{0.19}$ & $73.64_{0.19}$ & $81.33_{0.81}$ & $86.85_{0.72}$ \\
 & \textcolor{gray}{GPTQ sliced} & \textcolor{gray}{$53.58_{0.52}$} & \textcolor{gray}{$77.10_{0.91}$} & \textcolor{gray}{$73.61_{0.13}$} & \textcolor{gray}{$77.09_{0.34}$} & \textcolor{gray}{$68.59_{0.52}$} & \textcolor{gray}{$84.61_{0.30}$} & \textcolor{gray}{$73.05_{0.30}$} & \textcolor{gray}{$83.36_{0.42}$} & \textcolor{gray}{$88.61_{0.61}$} \\ \midrule
\multirow{6}{*}{3} & GPTQ & $52.05_{0.98}$ & $73.95_{0.99}$ & $72.41_{0.22}$ & $77.42_{0.68}$ & $70.80_{0.29}$ & $80.74_{0.70}$ & $70.66_{0.33}$ & $75.60_{0.52}$ & $82.85_{0.65}$ \\
 & AP & $49.78_{1.41}$ & $74.67_{2.03}$ & $69.95_{0.31}$ & $76.16_{0.43}$ & $64.83_{1.07}$ & $80.83_{1.23}$ & $69.43_{0.45}$ & $76.97_{1.31}$ & $83.69_{1.04}$ \\
 & AB & $53.07_{0.20}$ & $79.59_{0.10}$ & $70.71_{0.05}$ & $77.53_{0.08}$ & $66.61_{0.12}$ & $76.35_{0.04}$ & $69.11_{0.13}$ & $73.94_{0.40}$ & $81.29_{0.59}$ \\
 & MatQuant & $51.13_{0.64}$ & $75.45_{0.90}$ & $68.82_{0.22}$ & $75.24_{0.28}$ & $65.30_{0.66}$ & $77.04_{0.93}$ & $66.96_{0.15}$ & $74.53_{0.92}$ & $81.20_{0.81}$ \\
 & MatGPTQ & $51.48_{1.91}$ & $76.21_{1.73}$ & $70.93_{0.21}$ & $76.44_{0.42}$ & $66.98_{0.91}$ & $80.46_{1.55}$ & $69.08_{0.47}$ & $76.09_{0.38}$ & $82.77_{0.18}$ \\
 & \textcolor{gray}{GPTQ sliced} & \textcolor{gray}{$39.51_{1.02}$} & \textcolor{gray}{$58.46_{1.08}$} & \textcolor{gray}{$60.95_{0.22}$} & \textcolor{gray}{$71.11_{0.75}$} & \textcolor{gray}{$59.43_{0.35}$} & \textcolor{gray}{$54.44_{0.58}$} & \textcolor{gray}{$57.63_{0.40}$} & \textcolor{gray}{$48.98_{0.53}$} & \textcolor{gray}{$61.27_{0.70}$} \\
\midrule \midrule
\multirow{6}{*}{6} & GPTQ & $55.80_{0.24}$ & $80.13_{0.22}$ & $74.67_{0.08}$ & $77.31_{0.26}$ & $67.96_{0.23}$ & $87.04_{0.17}$ & $74.84_{0.11}$ & $81.33_{0.71}$ & $87.29_{0.47}$ \\
 & AP & $56.74_{0.45}$ & $80.97_{0.36}$ & $74.84_{0.11}$ & $77.46_{0.19}$ & $67.72_{0.54}$ & $87.45_{0.22}$ & $74.97_{0.13}$ & $81.48_{0.51}$ & $87.34_{0.36}$ \\
 & AB & $56.31_{0.52}$ & $80.61_{0.50}$ & $74.65_{0.18}$ & $77.62_{0.22}$ & $69.03_{0.66}$ & $87.29_{0.68}$ & $74.77_{0.08}$ & $80.65_{0.11}$ & $86.89_{0.18}$ \\
 & MatQuant & $55.78_{0.28}$ & $80.28_{0.11}$ & $74.74_{0.10}$ & $76.97_{0.14}$ & $67.73_{0.17}$ & $87.55_{0.33}$ & $74.95_{0.07}$ & $81.55_{0.68}$ & $87.22_{0.48}$ \\
 & MatGPTQ & $56.43_{0.64}$ & $80.01_{0.08}$ & $74.73_{0.02}$ & $77.57_{0.26}$ & $68.59_{0.49}$ & $87.34_{0.20}$ & $74.76_{0.11}$ & $80.59_{0.18}$ & $86.61_{0.30}$ \\
 & \textcolor{gray}{GPTQ sliced} & \textcolor{gray}{$55.97_{0.28}$} & \textcolor{gray}{$79.97_{0.19}$} & \textcolor{gray}{$74.70_{0.05}$} & \textcolor{gray}{$77.75_{0.40}$} & \textcolor{gray}{$68.27_{0.27}$} & \textcolor{gray}{$87.72_{0.17}$} & \textcolor{gray}{$74.83_{0.08}$} & \textcolor{gray}{$81.33_{0.74}$} & \textcolor{gray}{$87.29_{0.52}$} \\
\bottomrule
\end{tabular}
\label{tab:full_results_asym_qwen3_8b}
\end{table*}

\paragraph{Measuring variance.}
Every entry in Table~\ref{tab:main} is the mean over $3$ independently quantized models, each fitted from a different calibration seed and evaluated once. Because quantization is fitted on a small random sample of calibration data, we wanted to measure how far a score moves when that draw changes. Writing $s_i$ and $n_i$ for the sample standard deviation and number of calibration seeds of method $i$, the standard error of the difference between two entries is
\begin{equation}
\mathrm{SE}_{a-b}=\sqrt{\frac{s_a^2}{n_a}+\frac{s_b^2}{n_b}},
\end{equation}
and we mark an entry \textbf{bold} when a $95\%$ Welch interval, $t_{0.975,\nu}\,\mathrm{SE}_{a-b}$ with Welch-Satterthwaite degrees of freedom $\nu$, contains its gap to the best entry in the block, \underline{underlined} entries are indistinguishable from the full-precision model in the same sense. A bold entry is therefore one we cannot separate from the best given the seeds we have. 

\section{Detailed evaluations for the impact of loss weights $\lambda_R$}
\label{section:lambda-evaluation}

We compare a uniform weighting approach against configurations that highly weight either the lowest or the highest bit-width. Additionally, we compare increasing weights from low to high bit-widths and from high to low.  Tables~\ref{tab:ablation-llama} and~\ref{tab:ablation-qwen} present detailed evaluation results on the impact of the loss weights $(\lambda_3,\lambda_4,\lambda_8)$, for the asymmetric Llama3.1-8B-Instruct and Qwen3-8B models. For Llama, $(1,2,4)$ stays closest to the GPTQ baseline across all target bit-widths, with a worst-case gap of only $0.25$ points. For Qwen, $(1,1,1)$ performs better at 3 bits but loses performance at the higher bit-widths, however averaged over all target bit-widths, the 3-bit improvement slightly outweighs these losses. However, since we aim for a model that is balanced across bit-widths, i.e., with a consistently small gap to GPTQ at every precision, we select $(1,2,4)$ as the default configuration for both models.

Generally, we can summarize the $\lambda_r$ ablation as follows: the weights do matter. Focusing more strongly on higher bit-widths can worsen the lower-bit models, and vice versa. Therefore, depending on the target use case, $\lambda_r$ should be tuned according to which bit-widths are most important.

\begin{table*}[t]
\centering
\small
\setlength{\tabcolsep}{4pt}
\caption{Evaluation results for MatGPTQ on asymmetric Llama3.1-8B-Instruct for different $\lambda$ configurations, one triple applied to both the feedback and the scale-search channel. Weightings: $(x, y, z) \rightarrow (\lambda_3, \lambda_4, \lambda_8)$. IF-p and IF-i are the prompt- and instance-level strict IFEval accuracies; All weights four groups equally, the five common-sense tasks, MMLU, GSM8K and IFEval ($\uparrow$). The mean over three calibration seeds is reported.}
\label{tab:ablation-llama}
\begin{tabular}{ll rrrrr rrrr r}
\toprule
Bits & $\lambda$ & ARC-c & ARC-e & HellaS & PIQA & WinoG & GSM8K & MMLU & IF-p & IF-i & All \\
\midrule
\multirow{8}{*}{8} & (8, 1, 1) & 54.27 & 79.17 & 77.31 & 79.71 & 73.01 & 79.30 & 67.28 & 69.50 & 78.54 & 73.32 \\
 & (4, 2, 1) & 55.12 & 79.04 & 77.90 & 80.41 & 72.61 & 82.34 & 68.08 & 72.27 & 80.58 & 74.96 \\
 & (4, 4, 1) & 54.27 & 80.18 & 78.19 & 80.69 & 73.32 & 83.85 & 68.28 & 74.12 & 80.82 & 75.73 \\
 & (1, 1, 1) & 54.81 & 80.32 & 78.68 & 80.56 & 73.61 & 83.27 & 68.52 & 73.63 & 81.02 & 75.68 \\
 & (1, 2, 4) & 54.93 & 79.83 & 78.87 & 81.02 & 74.11 & 84.69 & 69.07 & 74.23 & 81.73 & 76.37 \\
 & (1, 4, 4) & 55.38 & 79.67 & 78.89 & 80.69 & 74.35 & 84.08 & 68.81 & 73.20 & 80.46 & 75.88 \\
 & (1, 1, 8) & 55.12 & 79.76 & 79.05 & 80.85 & 73.95 & 85.60 & 69.36 & 75.79 & 82.49 & 76.96 \\
\midrule
\multirow{8}{*}{4} & (8, 1, 1) & 54.01 & 79.25 & 76.55 & 79.98 & 72.77 & 77.86 & 66.55 & 66.91 & 76.14 & 72.11 \\
 & (4, 2, 1) & 55.12 & 80.51 & 77.38 & 80.79 & 73.40 & 79.76 & 67.38 & 70.43 & 78.78 & 73.79 \\
 & (4, 4, 1) & 53.84 & 79.50 & 77.62 & 80.52 & 73.88 & 80.29 & 67.92 & 70.98 & 79.14 & 74.09 \\
 & (1, 1, 1) & 53.70 & 78.82 & 78.06 & 80.81 & 73.69 & 81.68 & 67.91 & 71.47 & 79.70 & 74.55 \\
 & (1, 2, 4) & 53.96 & 78.65 & 78.42 & 80.76 & 73.62 & 82.84 & 68.33 & 73.42 & 81.03 & 75.37 \\
 & (1, 4, 4) & 54.78 & 79.17 & 78.35 & 80.96 & 73.80 & 82.71 & 68.03 & 74.12 & 81.06 & 75.44 \\
 & (1, 1, 8) & 54.78 & 79.38 & 78.00 & 81.45 & 74.11 & 81.43 & 68.05 & 72.46 & 80.70 & 74.90 \\
\midrule
\multirow{8}{*}{3} & (8, 1, 1) & 52.05 & 77.40 & 75.76 & 79.54 & 72.06 & 71.34 & 63.87 & 67.28 & 76.74 & 69.65 \\
 & (4, 2, 1) & 51.11 & 75.80 & 76.06 & 79.65 & 72.85 & 72.25 & 64.24 & 67.65 & 75.30 & 69.76 \\
 & (4, 4, 1) & 51.62 & 77.69 & 75.81 & 79.33 & 71.51 & 71.87 & 63.59 & 68.58 & 77.10 & 69.87 \\
 & (1, 1, 1) & 51.19 & 75.22 & 75.89 & 79.29 & 72.38 & 74.00 & 63.81 & 69.32 & 78.10 & 70.58 \\
 & (1, 2, 4) & 50.24 & 75.88 & 75.26 & 78.82 & 72.04 & 68.28 & 62.96 & 69.54 & 77.70 & 68.83 \\
 & (1, 4, 4) & 52.65 & 76.30 & 75.40 & 79.00 & 71.11 & 59.89 & 63.10 & 68.76 & 77.34 & 66.74 \\
 & (1, 1, 8) & 48.98 & 76.81 & 75.19 & 78.67 & 72.45 & 69.07 & 62.69 & 67.28 & 76.14 & 68.47 \\
\midrule \midrule
\multirow{8}{*}{6} & (8, 1, 1) & 54.86 & 79.67 & 77.38 & 79.87 & 72.69 & 79.00 & 67.02 & 70.61 & 79.02 & 73.43 \\
 & (4, 2, 1) & 54.95 & 79.50 & 77.80 & 80.47 & 73.09 & 82.18 & 67.81 & 70.98 & 79.14 & 74.55 \\
 & (4, 4, 1) & 54.69 & 79.84 & 78.18 & 80.96 & 74.03 & 83.40 & 68.29 & 71.72 & 79.74 & 75.24 \\
 & (1, 1, 1) & 54.72 & 79.91 & 78.63 & 80.67 & 73.66 & 82.84 & 68.49 & 73.26 & 80.66 & 75.45 \\
 & (1, 2, 4) & 54.88 & 79.60 & 78.82 & 81.01 & 74.14 & 84.82 & 68.93 & 73.94 & 81.61 & 76.30 \\
 & (1, 4, 4) & 54.95 & 79.59 & 78.92 & 80.74 & 74.03 & 84.15 & 68.84 & 73.57 & 80.46 & 75.91 \\
 & (1, 1, 8) & 55.03 & 79.46 & 78.90 & 81.18 & 73.72 & 84.69 & 69.04 & 73.38 & 80.70 & 76.10 \\
\bottomrule
\end{tabular}
\end{table*}

\begin{table*}[t]
\centering
\small
\setlength{\tabcolsep}{4pt}
\caption{Evaluation results for MatGPTQ on asymmetric Qwen3-8B for different $\lambda$ configurations, one triple applied to both the feedback and the scale-search channel. Weightings: $(x, y, z) \rightarrow (\lambda_3, \lambda_4, \lambda_8)$. IF-p and IF-i are the prompt- and instance-level strict IFEval accuracies; All weights four groups equally, the five common-sense tasks, MMLU, GSM8K and IFEval ($\uparrow$). The mean over three calibration seeds is reported.}
\label{tab:ablation-qwen}
\begin{tabular}{ll rrrrr rrrr r}
\toprule
Bits & $\lambda$ & ARC-c & ARC-e & HellaS & PIQA & WinoG & GSM8K & MMLU & IF-p & IF-i & All \\
\midrule
\multirow{8}{*}{8} & (8, 1, 1) & 56.83 & 79.12 & 72.81 & 77.26 & 68.51 & 86.66 & 72.94 & 80.41 & 86.21 & 78.45 \\
 & (4, 2, 1) & 54.44 & 78.70 & 73.45 & 77.48 & 69.93 & 86.96 & 73.65 & 78.19 & 84.77 & 78.22 \\
 & (4, 4, 1) & 56.46 & 80.06 & 73.66 & 77.71 & 69.17 & 86.53 & 74.14 & 80.47 & 86.41 & 78.88 \\
 & (1, 1, 1) & 56.57 & 79.86 & 74.11 & 77.76 & 68.82 & 86.75 & 74.58 & 80.55 & 86.57 & 79.08 \\
 & (1, 2, 4) & 56.03 & 80.13 & 74.68 & 77.53 & 68.14 & 87.36 & 74.73 & 81.76 & 87.53 & 79.51 \\
 & (1, 4, 4) & 55.80 & 79.84 & 74.58 & 77.58 & 68.59 & 86.96 & 74.54 & 82.26 & 87.77 & 79.45 \\
 & (1, 1, 8) & 56.23 & 80.43 & 74.81 & 77.75 & 67.64 & 87.41 & 75.15 & 81.15 & 87.53 & 79.57 \\
\midrule
\multirow{8}{*}{4} & (8, 1, 1) & 54.18 & 77.82 & 72.18 & 76.44 & 69.06 & 84.91 & 71.96 & 80.22 & 85.85 & 77.46 \\
 & (4, 2, 1) & 54.69 & 79.50 & 72.54 & 76.99 & 68.67 & 87.04 & 73.23 & 78.74 & 84.89 & 78.14 \\
 & (4, 4, 1) & 55.97 & 79.29 & 73.17 & 76.70 & 68.30 & 86.13 & 73.28 & 80.71 & 86.13 & 78.38 \\
 & (1, 1, 1) & 55.99 & 79.72 & 73.38 & 76.87 & 69.03 & 86.35 & 73.58 & 80.81 & 86.40 & 78.63 \\
 & (1, 2, 4) & 55.97 & 80.18 & 73.93 & 77.69 & 68.64 & 87.14 & 73.64 & 81.33 & 86.85 & 79.04 \\
 & (1, 4, 4) & 56.40 & 80.09 & 73.88 & 77.80 & 68.35 & 86.96 & 73.75 & 82.44 & 87.77 & 79.28 \\
 & (1, 1, 8) & 55.46 & 80.01 & 73.71 & 77.91 & 68.27 & 86.28 & 73.10 & 82.26 & 87.65 & 78.85 \\
\midrule
\multirow{8}{*}{3} & (8, 1, 1) & 53.41 & 77.02 & 72.28 & 76.28 & 67.25 & 83.62 & 71.34 & 76.52 & 83.45 & 76.05 \\
 & (4, 2, 1) & 53.84 & 79.67 & 72.62 & 77.26 & 67.25 & 82.56 & 70.60 & 79.67 & 85.61 & 76.48 \\
 & (4, 4, 1) & 53.07 & 78.17 & 72.49 & 76.82 & 67.64 & 82.79 & 71.00 & 75.54 & 82.97 & 75.67 \\
 & (1, 1, 1) & 53.52 & 79.72 & 72.06 & 76.99 & 67.23 & 82.74 & 70.80 & 77.01 & 83.67 & 75.95 \\
 & (1, 2, 4) & 51.48 & 76.21 & 70.93 & 76.44 & 66.98 & 80.46 & 69.08 & 76.09 & 82.77 & 74.35 \\
 & (1, 4, 4) & 49.66 & 73.02 & 69.87 & 75.68 & 66.54 & 81.20 & 69.19 & 73.94 & 81.53 & 73.77 \\
 & (1, 1, 8) & 50.68 & 75.21 & 70.60 & 77.58 & 65.67 & 78.17 & 68.52 & 75.42 & 82.85 & 73.44 \\
\midrule \midrule
\multirow{8}{*}{6} & (8, 1, 1) & 56.83 & 78.87 & 73.00 & 77.37 & 68.67 & 86.13 & 72.76 & 79.67 & 85.61 & 78.12 \\
 & (4, 2, 1) & 54.69 & 79.67 & 73.45 & 77.86 & 69.69 & 87.04 & 73.77 & 78.93 & 85.01 & 78.46 \\
 & (4, 4, 1) & 56.29 & 80.20 & 73.58 & 77.78 & 69.06 & 86.58 & 73.92 & 80.16 & 86.13 & 78.76 \\
 & (1, 1, 1) & 56.30 & 79.98 & 74.11 & 77.92 & 68.78 & 86.76 & 74.44 & 79.82 & 86.14 & 78.90 \\
 & (1, 2, 4) & 56.43 & 80.01 & 74.73 & 77.57 & 68.59 & 87.34 & 74.76 & 80.59 & 86.61 & 79.29 \\
 & (1, 4, 4) & 55.55 & 79.46 & 74.60 & 77.75 & 68.19 & 87.04 & 74.26 & 80.96 & 87.41 & 79.15 \\
 & (1, 1, 8) & 56.57 & 80.47 & 74.74 & 77.75 & 68.35 & 88.32 & 75.02 & 80.96 & 87.05 & 79.73 \\
\bottomrule
\end{tabular}
\end{table*}

\section{Detailed Kernel Performance Results}
\label{sec:kernel_details}

\begin{figure}[h]
    \centering
    \includegraphics[width=1\linewidth]{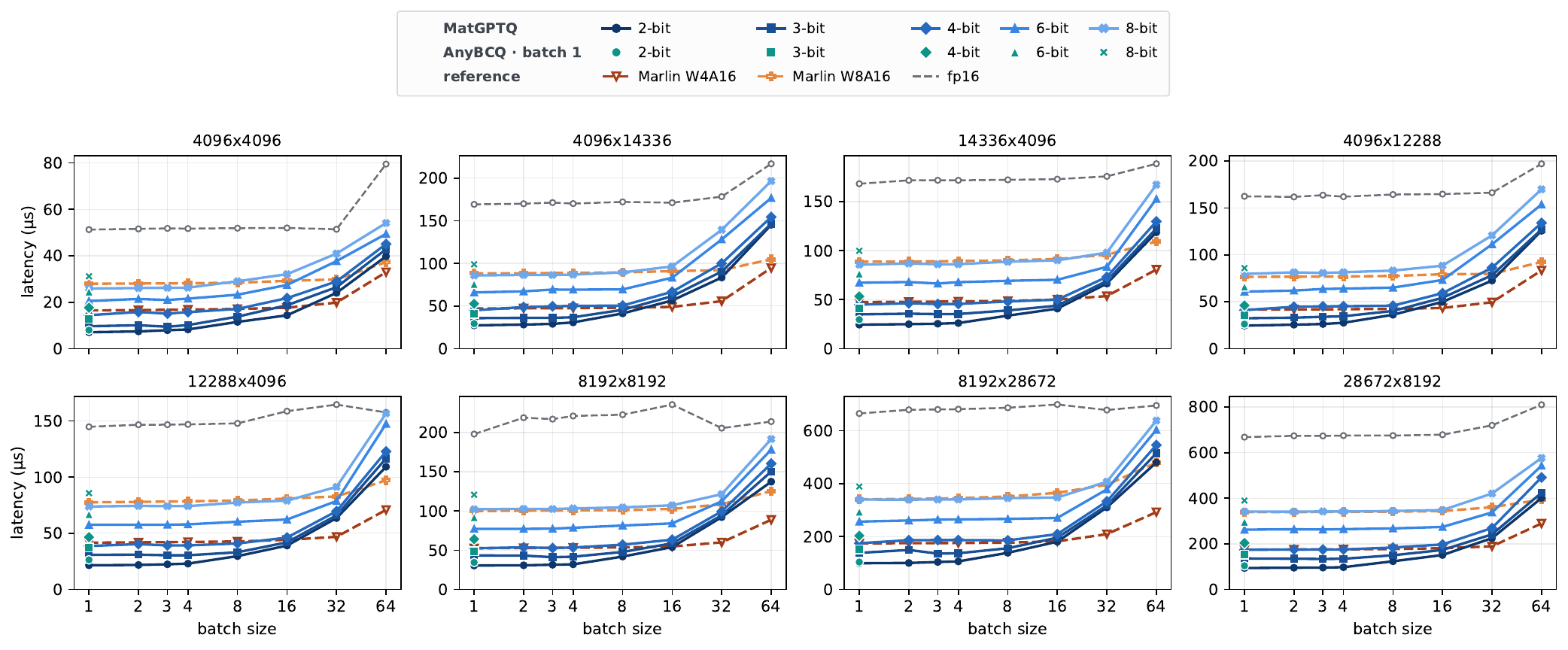}
    \caption{The figure shows the kernel latency compared to FP16 \texttt{torch.matmul} (cuBLAS). The kernels are evaluated at 2--8 bit precision for several representative matrix shapes across different batch sizes. AnyBCQ and Marlin W4A16/W8A16 are included as reference baselines. However, AnyBCQ results are only shown for batch size 1.}
    \label{fig:kernel-perf-latency}
\end{figure}

\begin{figure}[h]
  \centering

  \begin{minipage}[t]{0.49\linewidth}
    \centering
    \includegraphics[
      width=\linewidth,
      height=4.2cm,
      keepaspectratio
    ]{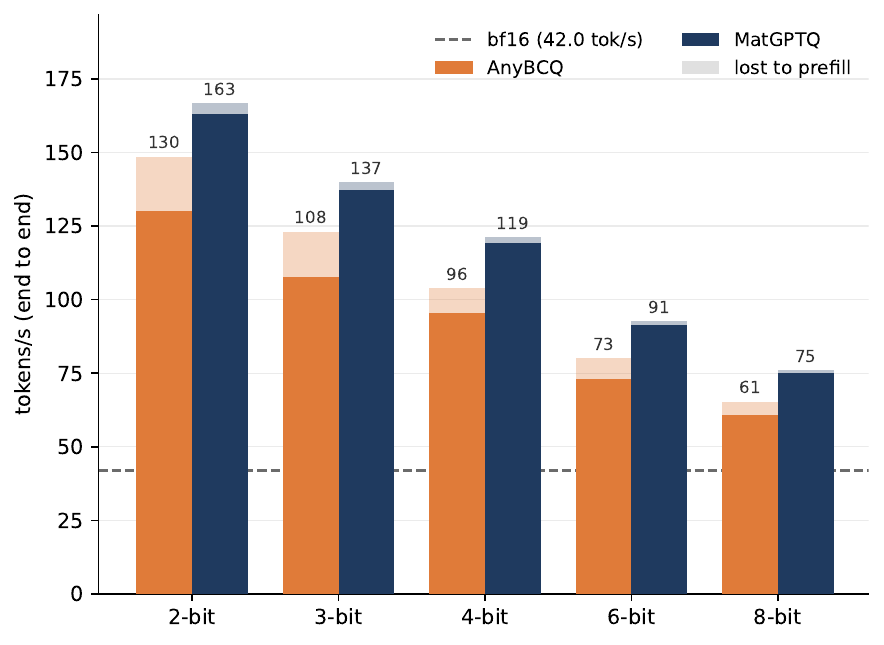}
    \captionof{figure}{End-to-end throughput: Transformers -- Llama3.1-8B-Instruct, Prompt/Decode 32/128.}
    \label{fig:e2e_short_tokps}
  \end{minipage}
  \hfill
  \begin{minipage}[t]{0.49\linewidth}
    \centering
    \includegraphics[
      width=\linewidth,
      height=4.2cm,
      keepaspectratio
    ]{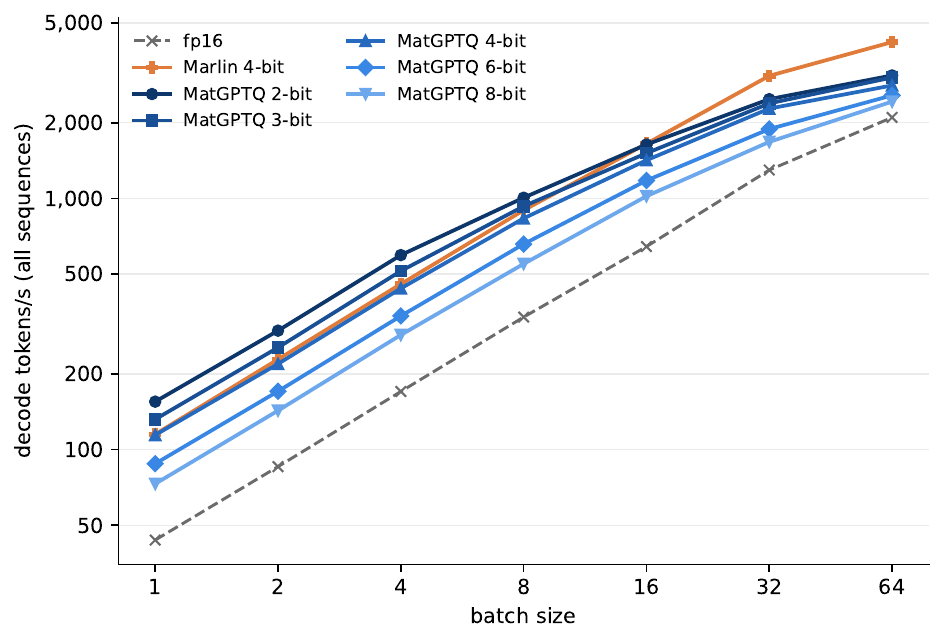}
    \captionof{figure}{End-to-end throughput: vLLM -- Llama3.1-8B-Instruct, Prompt/Decode 32/128.}
    \label{fig:e2e_vllm_tokps}
  \end{minipage}

\end{figure}

In Figures~\ref{fig:kernel-perf-latency}, \ref{fig:e2e_short_tokps}, and~\ref{fig:e2e_vllm_tokps}, we report the absolute latency and end-to-end throughput (tokens/s) for our kernel evaluations. These results complement the relative speedups reported in the main paper and provide a more detailed view of the kernel performance across bit-widths and batch sizes.

\begin{figure}[h]
    \centering
    \includegraphics[width=0.8\linewidth]{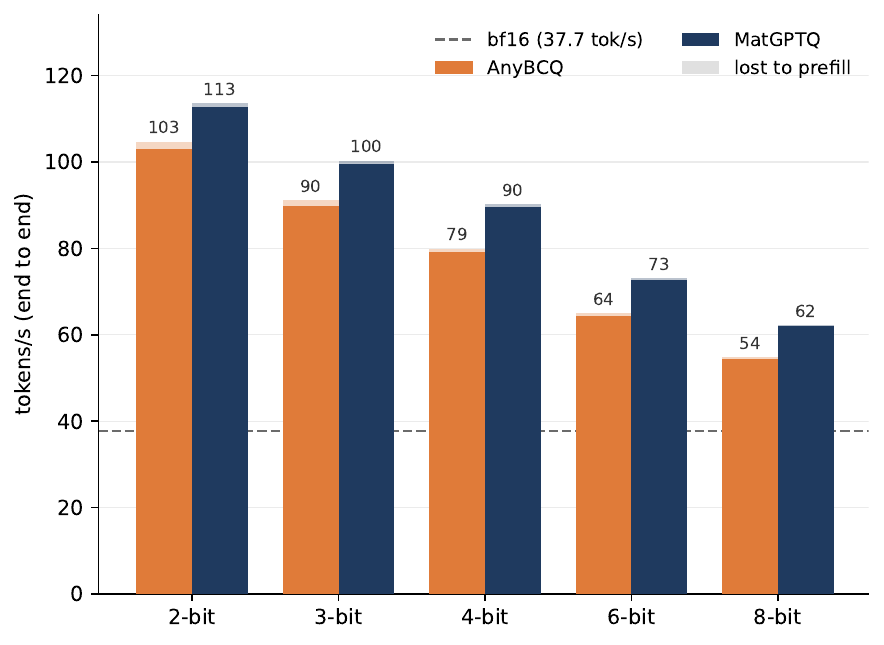}
    \caption{End-to-end throughput: Transformers -- Llama3.1-8B-Instruct, Prompt/Decode 256/1024.}
    \label{fig:e2e_longer}
\end{figure}

Additionally, we measured end-to-end speedups for longer prefill and decode workloads using our Transformer's integrations. As shown in Figure \ref{fig:e2e_longer}, MatGPTQ maintains its performance advantage. This can be attributed to our kernel matching baseline cuBLAS performance for longer prompt sequences, thereby at least matching AnyBCQ's dequantization+cuBLAS approach during prefill while providing faster decode performance. Overall, our system therefore provides better end-to-end system integration.

\end{document}